\documentclass[10pt,twocolumn,letterpaper]{article}

\usepackage[pagenumbers]{iccv}      %
\usepackage{makecell}
\usepackage{algorithm}
\usepackage{algorithmic}

\definecolor{iccvblue}{rgb}{0.21,0.49,0.74}
\usepackage[pagebackref,breaklinks,colorlinks,allcolors=iccvblue]{hyperref}

\usepackage{multirow}
\usepackage{xcolor}
\usepackage{colortbl}
\definecolor{baselinecolor}{gray}{.9}
\newcommand{\baseline}[1]{\cellcolor{baselinecolor}{#1}}

\newlength\savewidth\newcommand\shline{\noalign{\global\savewidth\arrayrulewidth
  \global\arrayrulewidth 1pt}\hline\noalign{\global\arrayrulewidth\savewidth}}

\newcommand{\tablestyle}[2]{\setlength{\tabcolsep}{#1}\renewcommand{\arraystretch}{#2}\centering\footnotesize}

\definecolor{commentcolor}{rgb}{0.2, 0.6, 0.2}
\definecolor{classcolor}{rgb}{0.7, 0.1, 0.2}
\definecolor{functioncolor}{rgb}{0.1, 0.1, 0.8}
\definecolor{keywordcolor}{rgb}{0.6, 0.2, 0.6}

\newcommand{\comment}[1]{\textcolor{commentcolor}{#1}}
\newcommand{\class}[1]{\textcolor{classcolor}{\textbf{#1}}}
\newcommand{\function}[1]{\textcolor{functioncolor}{\textbf{#1}}}
\newcommand{\keyword}[1]{\textcolor{keywordcolor}{\textbf{#1}}}

\newcommand{\modelname}{xAR\xspace}

\def\eg{\emph{e.g.}} 
\def\ie{\emph{i.e.}} 
 
 \def\vs{\emph{vs.}}

\newcommand{\tabref}[1]{Tab.~\ref{#1}}
\newcommand{\figref}[1]{Fig.~\ref{#1}}
\newcommand{\secref}[1]{Sec.~\ref{#1}}
\definecolor{order1}{RGB}{255, 0, 0}
\definecolor{order2}{RGB}{180, 210, 180}
\definecolor{order3}{RGB}{150, 170, 230}
\definecolor{order4}{RGB}{235, 222, 184}

\title{Beyond Next-Token: Next-X Prediction for Autoregressive Visual Generation}

\author{Sucheng Ren$^1$~~~ Qihang Yu$^2$~~~ Ju He$^2$~~~ Xiaohui Shen$^2$~~~ Alan Yuille$^1$~~~ Liang-Chieh Chen$^2$\\
$^1$Johns Hopkins University~~~~~$^2$ByteDance\\
\url{https://oliverrensu.github.io/project/xAR}
}

\begin{document}

\twocolumn[
{%
\maketitle\centering
\vspace{-20pt}
\includegraphics[width=0.916\linewidth]{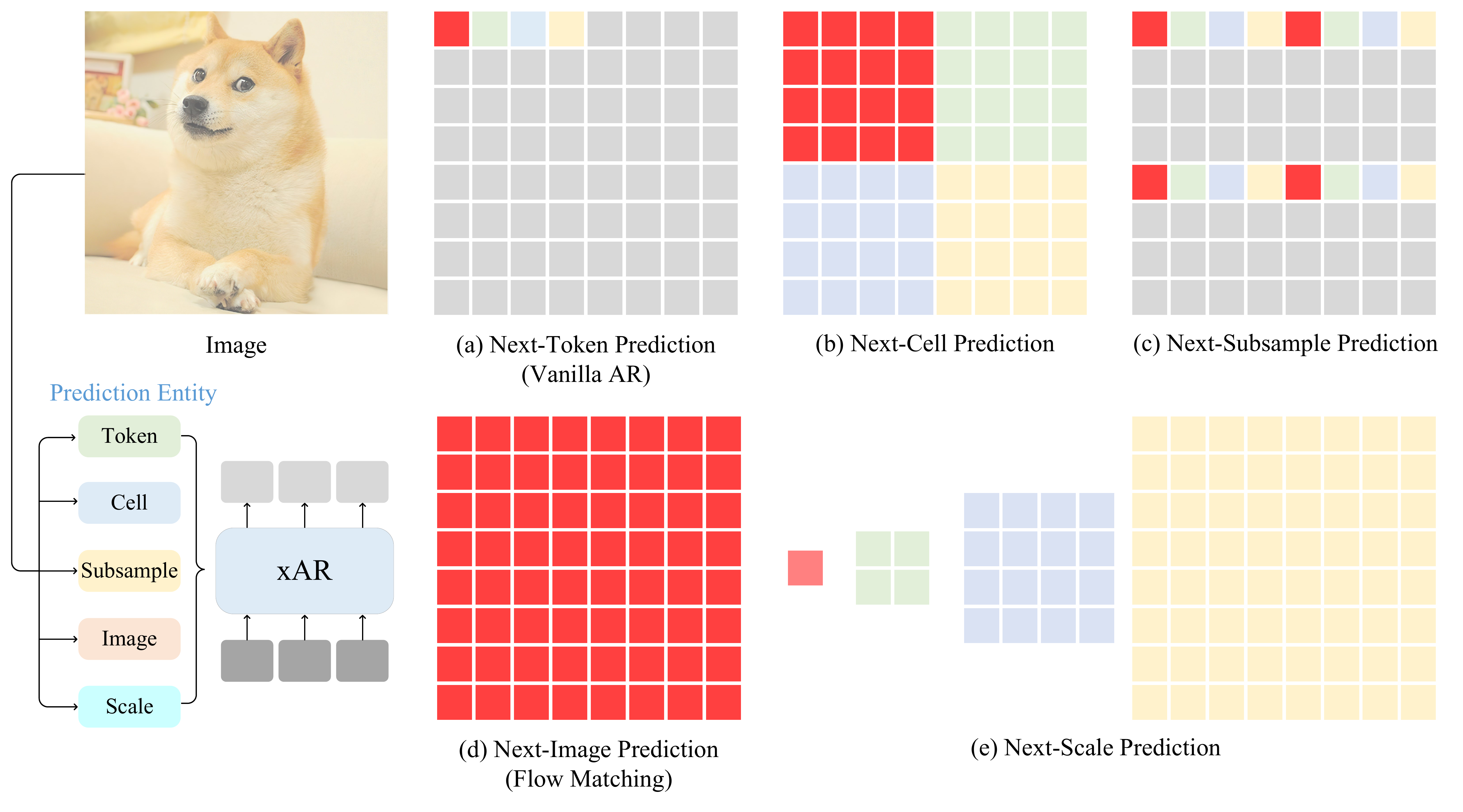}
  \vspace{-10pt}
  \captionof{figure}{
  \textbf{\modelname: Autoregressive (AR) Visual Generation with Next-X Prediction.}
  The proposed \modelname adopts a general next-X prediction framework, where X is a flexible prediction entity that can correspond to: (a) an individual image patch (as in vanilla AR~\cite{vqgan}), (b) a cell (a group of spatially contiguous tokens), (c) a subsample (a non-local grouping), (d) an entire image (as in flow-matching~\cite{lipman2022flow}), or (e) a scale (coarse-to-fine resolution, similar to VAR~\cite{var}).
  We use \textcolor{order1}{red}, \textcolor{order2}{green}, \textcolor{order3}{blue}, \textcolor{order4}{yellow} to illustrate the first four AR prediction steps for each entity example.
  The \textcolor{gray}{gray} tokens represent the remaining tokens.
  }
  \vspace{8pt}
  \label{fig:next_x}
}
]

\begin{abstract}
Autoregressive (AR) modeling, known for its next-token prediction paradigm, underpins state-of-the-art language and visual generative models. Traditionally, a ``token'' is treated as the smallest prediction unit, often a discrete symbol in language or a quantized patch in vision. However, the optimal token definition for 2D image structures remains an open question. Moreover, AR models suffer from exposure bias, where teacher forcing during training leads to error accumulation at inference.
In this paper, we propose \modelname, a generalized AR framework that extends the notion of a token to an entity X, which can represent an individual patch token, a cell (a $k\times k$ grouping of neighboring patches), a subsample (a non-local grouping of distant patches), a scale (coarse-to-fine resolution), or even a whole image. Additionally, we reformulate discrete token classification as \textbf{continuous entity regression}, leveraging flow-matching methods at each AR step. This approach conditions training on noisy entities instead of ground truth tokens, leading to Noisy Context Learning, which effectively alleviates exposure bias.
As a result, \modelname offers two key advantages: (1) it enables flexible prediction units that capture different contextual granularity and spatial structures, and (2) it mitigates exposure bias by avoiding reliance on teacher forcing.
On ImageNet-256 generation benchmark, our base model, \modelname-B (172M), outperforms DiT-XL/SiT-XL (675M) while achieving 20$\times$ faster inference. Meanwhile, \modelname-H sets a new state-of-the-art with an FID of 1.24, running 2.2$\times$ faster than the previous best-performing model without relying on vision foundation modules (\eg, DINOv2) or advanced guidance interval sampling.

\end{abstract}

\section{Introduction}
Autoregressive (AR) models have driven major advances in natural language processing (NLP) through next-token prediction, where each token is generated from its preceding tokens. This framework enables coherent, context-aware text generation, with landmark models like GPT-3~\cite{gpt3} and its successors~\cite{gpt4,chatgpt} setting new benchmarks across diverse NLP applications.

Building on the successes of AR modeling in NLP, researchers have extended this framework to computer vision, particularly for high-fidelity image generation~\cite{vqgan,parti,llamagen,mar,yu2024randomized}. In these approaches, image patches are discretized into tokens~\cite{vqvae} and reshaped into 1D sequences, allowing AR models to predict each token sequentially. However, unlike language, where tokens correspond to semantically meaningful units such as words, vision lacks a universally agreed-upon token definition. This naturally raises the question: 
\emph{How can ``next-token prediction'' be generalized to ``next-X prediction,'' and what constitutes the most suitable X for image generation?}

Additionally, beyond token design, traditional AR models rely on teacher forcing~\cite{williams1989learning} during training, where ground truth tokens are provided at each step instead of the model’s own predictions. While this stabilizes training, it introduces exposure bias~\cite{ranzato2016sequence}, since the model is never exposed to potential errors.
Consequently, during inference, without ground truth guidance, errors accumulate over time, leading to cascading errors and context drift as the model conditions solely on its past predictions.

To address these challenges, we propose \modelname, a general next-X prediction framework that reformulates discrete token \textit{classification} (conditioned on all preceding discrete \textit{ground truth} tokens) into a continuous entity \textit{regression} problem conditioned on all previous \textit{noisy} entities. The regression process is guided by flow-matching~\cite{liu2022flow,lipman2022flow} at each AR step. As illustrated in~\figref{fig:next_x}, within this framework, X serves as a flexible representation that can correspond to an individual patch token, a cell (a group of surrounding tokens), a subsample (a non-local grouping), a scale (coarse-to-fine resolution), or even an entire image.

Unlike teacher forcing~\cite{williams1989learning}, which always provides ground truth inputs, \modelname deliberately exposes the model to noisy contexts during training, allowing it to learn from imperfect, corrupted, or partially inaccurate conditions. We refer to this approach as Noisy Context Learning (NCL), a reformulation that reduces reliance on ground truth inputs, improving robustness and mitigating exposure bias~\cite{ranzato2016sequence} by enabling the model to generalize better during inference.

We demonstrate the effectiveness of \modelname on the challenging ImageNet generation benchmark~\cite{deng2009imagenet}. Through systematic experimentation with different X configurations, we find that \textbf{\textit{next-cell}} prediction—where neighboring tokens are grouped into moderately sized cells (\eg, 8$\times$8 tokens)—yields the best performance by capturing richer spatial-semantic relationships. Leveraging both next-cell prediction and Noisy Context Learning, our base model \modelname-B (172M) outperforms the large DiT-XL~\cite{dit} and SiT-XL~\cite{sit} (675M) while achieving 20$\times$ faster inference. Additionally, our largest model, \modelname-H (1.1B), sets a new state-of-the-art with an FID of 1.24 and runs 2.2$\times$ faster than the previous best-performing model~\cite{yu2024representation} on ImageNet-256~\cite{deng2009imagenet}, without relying on vision foundation models (\eg, DINOv2~\cite{dinov2}) or extra guidance interval sampling~\cite{guidance}.

\begin{figure}[t!]
    \centering
    \vspace{-10pt}
    \includegraphics[width=0.88\linewidth]{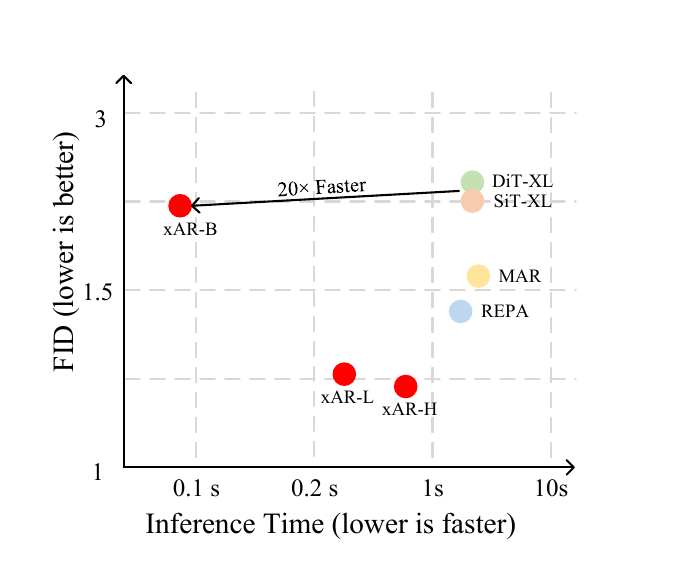}
    \vspace{-6pt}
    \caption{
    \textbf{ImageNet-256 Results.}
    Our base model, \modelname-B, outperforms DiT-XL~\cite{dit} and SiT-XL~\cite{sit} while achieving 20$\times$ faster inference, and our largest model, \modelname-H, establishes a new state-of-the-art with an FID of 1.24 on ImageNet-256.
    }
    
    \label{fig:teaser}
\end{figure}  

\section{Related Work}
\noindent\textbf{AR Modeling in NLP.}
Autoregressive language models~\cite{llama,gpt,gpt3,gpt4,chatgpt} have driven significant progress toward general-purpose AI. Their core principle is simple yet powerful: predicting the next token based on preceding context. This approach has demonstrated impressive scalability, guided by scaling laws, and adaptability, enabling zero-shot generalization. These strengths have extended AR modeling beyond traditional language tasks, influencing a wide range of modalities.

\noindent\textbf{AR Modeling in Vision.}
Inspired by the success of AR modeling in NLP, researchers have explored its application in vision~\cite{var,llamagen,igpt,pixelcnn,parti,yu2024randomized,vqvae2}. A pioneering effort in this direction was PixelCNN~\cite{pixelcnn}, which factorized the joint pixel distribution into a product of conditionals, enabling the model to learn complex image distributions. This idea was further refined in PixelRNN~\cite{pixelrnn}, which incorporated recurrent layers to capture richer context in both horizontal and vertical directions. iGPT~\cite{igpt} extended this pixel-level approach by leveraging Transformers~\cite{vaswani2017attention} for next-pixel prediction.
Beyond next-pixel modeling, AR methods have shifted toward more abstract token representations. VQ-VAE~\cite{vqvae} introduced discrete latent codes that could be modeled autoregressively, offering a compressed yet expressive representation of images. Later models like Parti~\cite{parti} and LlamaGen~\cite{llamagen} combined these learned tokens with Transformer-based architectures to generate high-fidelity images while maintaining scalable training.
Recently, MAR~\cite{mar} introduced a diffusion-based approach~\cite{song2019generative,diff2} to model per-token probability distributions in a continuous space, replacing categorical cross-entropy with a diffusion loss. VAR~\cite{var} extended next-token prediction to a coarse-to-fine scale prediction paradigm, progressively refining image details.
Our work unifies these approaches under a general next-X prediction framework, where X can flexibly represent tokens, scales, or our newly introduced cells, providing a more flexible and generalizable formulation for autoregressive visual modeling.

\noindent\textbf{Diffusion and Flow Matching.}
Beyond autoregressive modeling, diffusion~\cite{song2019generative,diff2,diff3} and flow matching~\cite{liu2022flow,lipman2022flow,sd3} have surpassed Generative Adversarial Networks (GANs)~\cite{gan,stylegan-xl} by employing multi-step denoising.
Latent Diffusion Models (LDMs)~\cite{ldm} improve speed and scalability by operating in a compressed latent space~\cite{vae} instead of raw pixels. Building on this, DiT~\cite{dit} and U-ViT~\cite{uvit} replace the traditional convolution-based U-Net~\cite{unet} with Transformers~\cite{vaswani2017attention} in latent space, further enhancing performance. Simple Diffusion~\cite{diff1,hoogeboom2024simpler} introduces a streamlined approach for scaling pixel-space diffusion models to high-resolution outputs, while DiMR~\cite{liu2024alleviating} progressively refines features across multiple scales, improving detail from low to high resolution.
In parallel, flow matching~\cite{liu2022flow,lipman2022flow} reformulates the generative process by directly mapping data distributions to a standard normal distribution, simplifying the transition from noise to structured data. SiT~\cite{sit} builds on this by integrating flow matching into DiT’s Transformer backbone for more efficient distribution alignment. Extending this approach, SD3~\cite{sd3} introduces a Transformer-based architecture that leverages flow matching for text-to-image generation. REPA~\cite{yu2024representation} refines denoising by aligning noisy intermediate states with clean image embeddings extracted from pretrained visual encoders~\cite{dinov2}.

\section{Method}
\label{sec:method}
In this section, we first provide an overview of autoregressive modeling with the next-token prediction paradigm in~\secref{sec:ar}, followed by our proposed xAR framework with next-X prediction and Noisy Context Learning in~\secref{sec:xar}.

\subsection{Preliminary: Next-Token Prediction}
\label{sec:ar}
Autoregressive modeling with next-token prediction is a fundamental approach in language modeling where the joint
probability of a token sequence is factorized into a product
of conditional probabilities. Formally, given a sequence \( \boldsymbol{x} = \{x_{1}, x_{2}, \dots, x_{N}\} \),
the model estimates
\begin{equation}
  P(\boldsymbol{x}) = \prod_{n=1}^{N} P \bigl( x_{n} \mid x_{1}, x_{2}, \dots, x_{n-1} \bigr).
\end{equation}
In practice, an autoregressive language model predicts the next token \( x_{n} \) through token classification, conditioned on all preceding tokens \( \{x_{1}, x_{2}, \dots, x_{n-1}\} \). This process proceeds sequentially from left to right (\ie, $n=\{1, \dots, N\}$) until the full sequence is generated.
For visual generation, a VQ tokenizer~\cite{vqvae,vqgan}  discretizes an image into a sequence of tokens. An autoregressive visual generation model then follows the next-token prediction paradigm, sequentially predicting tokens through classification conditioned on previously generated tokens. However, directly applying the next-token prediction paradigm to visual generation introduces several challenges:

\noindent\textbf{Information Density.}
In NLP, each token (\eg, a word) carries rich semantic meaning. In contrast, visual tokens typically represent small image patches, which may not be as semantically meaningful in isolation. A single patch can contain fragments of different objects or textures, making it difficult for the model to infer meaningful relationships between consecutive patches. Additionally, the quantization process in VQ-VAE~\cite{vqvae} can discard fine details, leading to lower-quality reconstructions. As a result, even if the model predicts the next token correctly, the generated image may still appear blurry or lack detail.

\noindent\textbf{Accumulated Errors.}
Teacher forcing~\cite{williams1989learning}, a common training strategy, feeds the model ground truth tokens to stabilize learning. However, this reliance on perfect context causes exposure bias~\cite{acc1,acc2}—the model never learns to recover from its potential mistakes. During inference, when it must condition on its own predictions, small errors can accumulate over time, leading to compounding artifacts and degraded output quality. 

To address these challenges, we extend next-token prediction to \textit{next-X prediction}, transitioning from traditional AR to \modelname. This is accomplished by introducing a more expressive prediction entity X and training the model with noisy entities for improved robustness.

\subsection{The Proposed \modelname}
\label{sec:xar}
We introduce \modelname, which consists of two key components: next-X prediction (\secref{sec:next_x}) and Noisy Context Learning  (\secref{sec:self_correct}). We first detail each component, then describe the inference strategy  (\secref{sec:inference}), followed by a discussion on how \modelname enhances visual generation (\secref{sec:xar_discussion}).

\subsubsection{Next-X Prediction}
\label{sec:next_x}
Given an image, we use an off-the-shelf VAE~\cite{vae} (instead of VQ-VAE~\cite{vqvae} to avoid quantization loss) to convert it into a continuous latent $I \in \mathcal{R}^{\frac{H}{f}\times \frac{W}{f}\times C}$, where $H$ and $W$ denote image height and width, $f$ is the downsampling rate (we use $f=16$~\cite{mar}), and $C$ represents the number of channels.
We then construct a sequence of prediction entities $\boldsymbol{X}=\{X_1, X_2, \dots, X_N\}$ based on $I$.
Each $X_i$ is a flexible entity that can represent 
an individual token (an image patch), a cell (a group of surrounding tokens),
a subsample (a non-local grouping), a scale (coarse-to-fine
resolution), or even an entire image.
We outline common choices for X below and refer readers to~\figref{fig:next_x} for visualization and Algorithm~\ref{algo:pseduo_code} for a PyTorch pseudo-code implementation.

\noindent\textbf{Individual Patch Token (\figref{fig:next_x}~(a)).}
When $X_i$ corresponds to a single image patch, \modelname reduces to standard AR modeling, where each token is predicted sequentially.

\noindent\textbf{Cell (\figref{fig:next_x}~(b)).}
The image is divided into an $m\times m$ grid, where each cell has $k \times k$ spatially adjacent tokens\footnote{We also experimented with rectangular cells (\eg, cells with shape $k/2 \times 2k$ or $2k \times k/2$), but observed no significant difference compared to squared cells. Thus, we adopt the simpler squared cell design.}.

\noindent\textbf{Subsample (\figref{fig:next_x}~(c)).}
Entities are created by spatially and uniformly subsampling the image grid~\cite{vqgan}.

\noindent\textbf{Entire Image (\figref{fig:next_x}~(d)).}
As an extreme case, all tokens are grouped into a single entity, \ie, $X = X_1 = I$, transforming \modelname into a flow matching method~\cite{liu2022flow,lipman2022flow}.

\noindent\textbf{Scale (\figref{fig:next_x}~(e)).}
A multi-scale hierarchical representation is constructed, similar to VAR~\cite{var}. Given any scale design $\{s_1, \dots, s_N\}$, we define $X_i = \mathrm{resize}(I, s_i)$, where $\mathrm{resize}$ refers to resizing the latent $I$ to the target scale $s_i$.
By default, we set $X_N = I \in \mathcal{R}^{\frac{H}{f}\times \frac{W}{f}\times C}$ (\ie, $s_N=\frac{H}{f}$), and define
$X_i = \mathrm{resize}(I, \frac{H}{f} \cdot \frac{1}{2^{N-i}})$ (\ie, $s_i=\frac{H}{f} \cdot \frac{1}{2^{N-i}}$)
which progressively refines predictions from coarse to fine scales. Unlike VAR~\cite{var}, our approach generalizes next-scale prediction to any scale configuration and does not require a specially designed multi-scale VQGAN tokenizer.

\noindent\textbf{Default Choice of X.}
Extensive ablation studies in~\secref{sec:ablation} show that cell (with a size of 8$\times$8 tokens) achieves the best performance among all X designs. Therefore, unless specified otherwise, \modelname adopts 8$\times$8 cells as the default X.

\begin{algorithm}[t]
\caption{PyTorch Pseudo-Code for General Entity X
}
\label{algo:pseduo_code}
\begin{algorithmic}[0]
\footnotesize
\STATE \keyword{from} einops \keyword{import} rearrange
\STATE \keyword{import} torch
\STATE \keyword{import} torch.nn.functional \keyword{as} F
\STATE \class{class} xAR(\keyword{nn.Module}): 
\STATE \quad \quad \comment{\# Construct a sequence of entities based on the input latent.}
\STATE \quad \quad \comment{\# Input: A continuous latent with shape (b, c, h, w).}
\STATE \quad \quad \comment{\# Return: A sequence of entities with shape (b, s, c).}
\STATE \quad \function{def} latent2token(self, latent): 
\STATE \quad \quad return latent.flatten(2).permute(0,2,1)
\STATE \quad \function{def} latent2cell(self, latent, k): 
\STATE \quad \quad \comment{\# k: Group $k\times k$ spatially neighboring tokens into one cell.}
\STATE \quad \quad return rearrange(latent, "b c (h k1) (w k2) -\verb|>| b (h w k1 k2) c", k1=k, k2=k)
\STATE \quad \function{def} latent2subsample(self, latent, distance): 
\STATE \quad \quad \comment{\# distance: Group tokens based on evenly spaced distances.}
\STATE \quad \quad return rearrange(latent, "b c (d1 h) (d2 w) -\verb|>| b (h w d1 d2) c", d1=distance, d2=distance)
\STATE \quad \function{def} latent2scale(self, latent, scales): 
\STATE \quad \quad \comment{\# scales: A sequence of scale design.}
\STATE \quad \quad entities = [F.interpolate(latent, (i,i)).flatten(2).permute(0,2,1) for i in scales]
\STATE \quad \quad entities = torch.cat(entities, dim=1)
\STATE \quad \quad return entities
\end{algorithmic}
\end{algorithm}

\begin{figure*}[t]
    \centering
    \vspace{-14pt}
    \includegraphics[width=\linewidth]{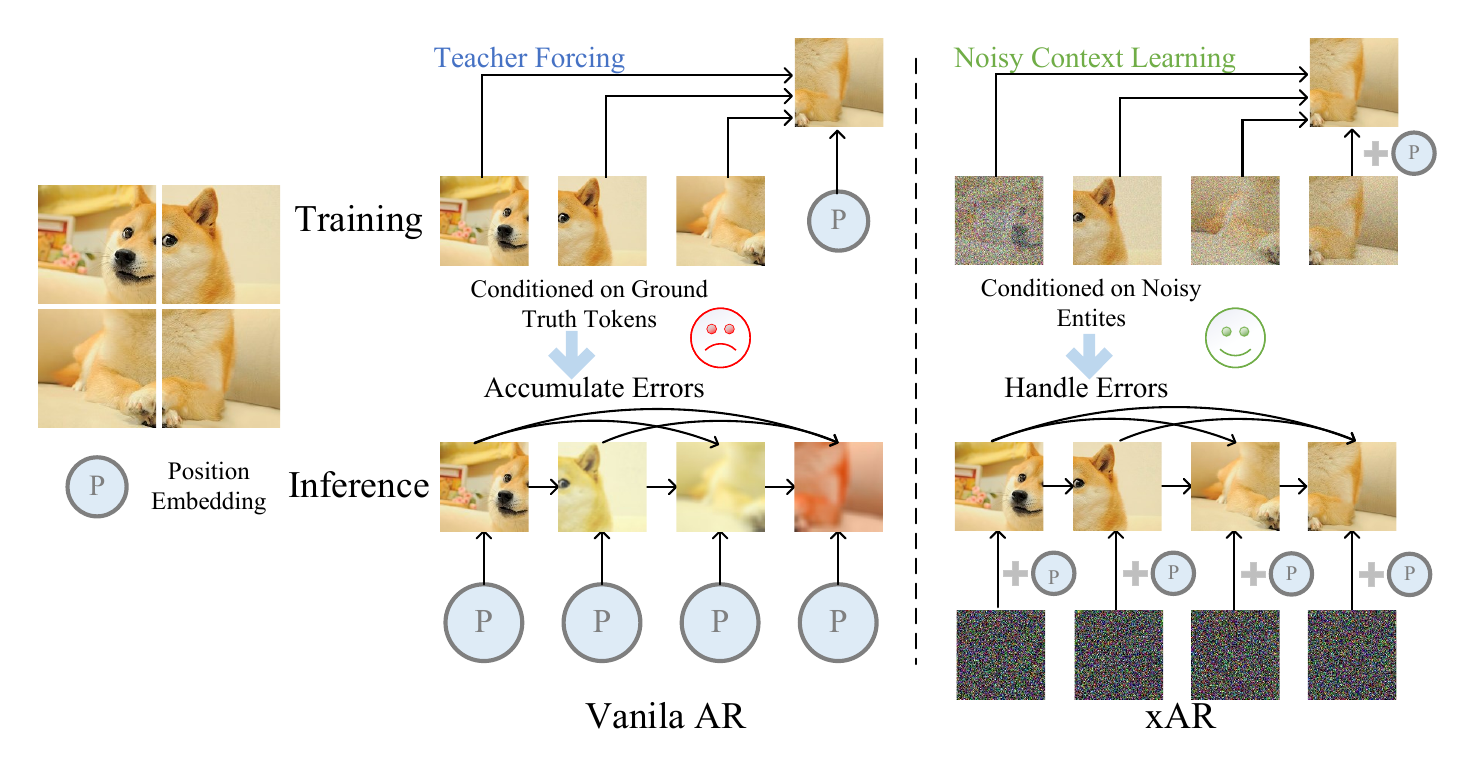}
    \vspace{-26pt}
    \caption{
    \textbf{Conditioning Mechanism Comparison between Vanilla AR \vs \ \modelname.}
    During training, vanilla AR conditions on all preceding ground truth tokens (\ie, Teacher Forcing), whereas \modelname conditions on all previous noisy entities, each with different noises (\ie, Noisy Context Learning).
    At inference, vanilla AR suffers from exposure bias, as errors accumulate over AR steps due to its exclusive training on ground truth tokens, leaving it unprepared for imperfect predictions. In contrast, \modelname, trained to handle noisy inputs, reduces reliance on ground truth signals and improves robustness to prediction errors.
    }
    \label{fig:method}
\end{figure*}

\subsubsection{Noisy Context Learning}
\label{sec:self_correct}

\modelname transitions the paradigm from ``discrete token classification'' (conditioned on all preceding \emph{ground truth} tokens) to ``continuous entity regression'' (conditioned on all previous \emph{noisy} entities).
Specifically, unlike traditional AR modeling, which directly classifies \(X_n\) based on all preceding ground truth entities \(\{X_1, \dots, X_{n-1}\}\), \modelname predicts \(X_n\) by minimizing a regression loss derived from flow matching~\cite{lipman2022flow,liu2022flow}, conditioned on all previous noisy entities.

During training, we randomly sample \(n\) noise time steps 
\(\{t_1, \dots, t_n\} \subset [0, 1]\), and draw \(n\) noise samples \(\{\epsilon_1, \dots, \epsilon_n\}\) from the source Gaussian noise distribution.
Specifically, at the $n$-th AR step, the noise samples are drawn as \(\epsilon_n \sim \mathcal{N}(0, I)\), where $\epsilon_n$ and $X_n$ share the same shape~\cite{ldm}.
We construct the interpolated input \(F_n^{t_n}\) as:
\begin{equation}
    F_n^{t_n} = \bigl(1 - t_n\bigr)  X_n + t_n \epsilon_n.
\end{equation}
Note that in $F$, the superscript denotes the flow-matching noise time step, while the subscript represents the AR time step.
We then define the velocity \(V_n^{t_n}\) as:
\begin{equation}
\begin{split}
    V_n^{t_n} &= \frac{dF_n^{t_n}}{dt_n} \\
              &=  \epsilon_n - X_n,
\end{split}
\end{equation}
where \(V_n^{t_n}\) represents the directional flow from \(F_n^{t_n}\) toward \(X_n\), guiding the transformation from the source to the target distribution.

The model is trained to predict the velocity \(V_n^{t_n}\) using all preceding and current noisy entities \(\{F_1^{t_1}, \dots, F_{n}^{t_{n}}\}\):
\begin{equation}
    \mathcal{L} = \sum_{n=1}^N \Bigl\| \mathrm{xAR}\bigl(\{F_1^{t_1}, \dots, F_{n}^{t_{n}}\}, t_n; \theta\bigr) - V_n^{t_n} \Bigr\|^2,
\end{equation}
where \(\mathrm{xAR}\) denotes our \modelname model parameterized by \(\theta\).

We refer to this scheme as Noisy Context Learning (NCL), where the model is trained by conditioning on all previous noisy entities rather than perfect ground truth inputs.
This effectively reduces reliance on clean training signals, improving robustness and mitigating exposure bias~\cite{ranzato2016sequence}.
\figref{fig:method} (Training) provides an illustration of NCL.
Notably, when sampling the time steps \(\{t_1, \dots, t_n\} \subset [0, 1]\), no constraints are imposed (\eg, we do not enforce $t_1>t_2$), allowing the model to experience varying degrees of noise in preceding entities, strengthening its adaptability during inference.

\subsubsection{Inference Scheme}
\label{sec:inference}
\modelname performs autoregressive prediction at the level of entity X. Since ``cell'' is the default choice for X, we use it as a concrete example.
As illustrated in \figref{fig:method} (Inference), \modelname begins by predicting an initial cell $\hat{X}_1$ from a Gaussian noise sample $\epsilon_1 \sim \mathcal{N}(0, I)$ (where $\epsilon_1$ has the same shape as $\hat{X}_1$) via flow matching~\cite{lipman2022flow,liu2022flow}.
Conditioned on the clean estimate $\hat{X}_1$, \modelname generates the next cell $\hat{X}_2$ from another Gaussian noise sample $\epsilon_2$.
This process continues autoregressively, where at the 
$i$-th AR step, the model predicts the next cell $\hat{X}_i$ based on all previously generated clean cells $\{\hat{X}_1, \cdots, \hat{X}_{i-1}\}$ and the newly drawn Gaussian noise sample $\epsilon_i$.
This iterative approach progressively refines the image, ensuring structured and context-aware generation at the cell level.

\subsubsection{Discussion}
\label{sec:xar_discussion}
As discussed in~\secref{sec:ar}, traditional AR modeling for visual generation faces two key challenges: information density and accumulated errors. The proposed \modelname is designed to address these limitations.

\noindent\textbf{Semantic-Rich Prediction Entity.} A cell (\ie, a $k\times k$ grouping of spatially contiguous tokens) aggregates neighboring tokens, effectively capturing both local structures (\eg, edges, textures) and regional contexts (\eg, small objects or parts of larger objects).
This leads to richer semantic representations compared to single-token predictions. By modeling relationships within the cell, the model learns to generate coherent local and regional features, shifting from isolated token-level predictions to holistic patterns.
Additionally, predicting a cell rather than an individual token allows the model to reason at a higher abstraction level, akin to how NLP models predict words instead of characters. The larger receptive field per prediction step contributes more semantic information, bridging the gap between low-level visual patches and high-level semantics.

\noindent\textbf{Robustness to Previous Prediction Errors.} 
The Noisy Context Learning (NCL) strategy trains the model on noisy entities instead of perfect ground truth inputs, reducing over-reliance on pristine contexts. This alignment between training and inference distributions enhances the model’s ability to handle errors in self-generated predictions.
By conditioning on imperfect contexts, \modelname learns to tolerate minor inaccuracies, preventing small errors from compounding into cascading errors. Additionally, exposure to noisy inputs encourages smoother representation learning, leading to more stable and consistent generations.

\section{Experimental Results}
In this section, we present the main results in~\secref{sec:main}, followed by ablation studies on key design choices in~\secref{sec:ablation}.

\begin{table*}[t]
\renewcommand\arraystretch{1.05}
\centering
\setlength{\tabcolsep}{2.5mm}{}
\begin{tabular}{l|l|c|cc|cc}
type & model     & \#params      & FID$\downarrow$ & IS$\uparrow$ & Precision$\uparrow$ & Recall$\uparrow$ \\
\shline
GAN& BigGAN~\cite{biggan} & 112M & 6.95  & 224.5       & 0.89 & 0.38     \\
GAN& GigaGAN~\cite{gigagan}  & 569M      & 3.45  & 225.5       & 0.84 & 0.61\\  
GAN& StyleGan-XL~\cite{stylegan-xl} & 166M & 2.30  & 265.1       & 0.78 & 0.53  \\
\hline
Diffusion& ADM~\cite{adm}    & 554M      & 10.94 & 101.0        & 0.69 & 0.63\\
Diffusion& LDM-4-G~\cite{ldm}   & 400M  & 3.60  & 247.7       & -  & -     \\
Diffusion & Simple-Diffusion~\cite{diff1} & 2B & 2.44 & 256.3 & - & - \\
Diffusion& DiT-XL/2~\cite{dit} & 675M     & 2.27  & 278.2       & 0.83 & 0.57     \\
Diffusion&L-DiT-3B~\cite{dit-github}  & 3.0B    & 2.10  & 304.4       & 0.82 & 0.60    \\
Diffusion&DiMR-G/2R~\cite{liu2024alleviating} &1.1B& 1.63& 292.5& 0.79 &0.63 \\
Diffusion & MDTv2-XL/2~\cite{gao2023mdtv2} & 676M & 1.58 & 314.7 & 0.79 & 0.65\\
Diffusion & CausalFusion-H$^\dag$~\cite{deng2024causal} & 1B & 1.57 & - & - & - \\
\hline
Flow-Matching & SiT-XL/2~\cite{sit} & 675M & 2.06 & 277.5 & 0.83 & 0.59 \\
Flow-Matching&REPA~\cite{yu2024representation} &675M& 1.80 & 284.0 &0.81 &0.61\\    
Flow-Matching&REPA$^\dag$~\cite{yu2024representation}& 675M& 1.42&  305.7& 0.80& 0.65 \\
\hline
Mask.& MaskGIT~\cite{maskgit}  & 227M   & 6.18  & 182.1        & 0.80 & 0.51 \\
Mask. & TiTok-S-128~\cite{yu2024image} & 287M & 1.97 & 281.8 & - & - \\
Mask. & MAGVIT-v2~\cite{yu2024language} & 307M & 1.78 & 319.4 & - & - \\ 
Mask. & MaskBit~\cite{weber2024maskbit} & 305M & 1.52 & 328.6 & - & - \\
\hline
AR& VQVAE-2~\cite{vqvae2} & 13.5B    & 31.11           & $\sim$45     & 0.36           & 0.57          \\
AR& VQGAN~\cite{vqgan}& 227M  & 18.65 & 80.4         & 0.78 & 0.26   \\
AR& VQGAN~\cite{vqgan}   & 1.4B     & 15.78 & 74.3   & -  & -     \\
AR&RQTran.~\cite{rq}     & 3.8B    & 7.55  & 134.0  & -  & -    \\
AR& ViTVQ~\cite{vit-vqgan} & 1.7B  & 4.17  & 175.1  & -  & -    \\
AR & DART-AR~\cite{gu2025dart} & 812M & 3.98 & 256.8 & - & - \\
AR & MonoFormer~\cite{zhao2024monoformer} & 1.1B & 2.57 & 272.6 & 0.84 & 0.56\\
AR & Open-MAGVIT2-XL~\cite{luo2024open} & 1.5B & 2.33 & 271.8 & 0.84 & 0.54\\
AR&LlamaGen-3B~\cite{llamagen}  &3.1B& 2.18& 263.3 &0.81& 0.58\\
AR & FlowAR-H~\cite{flowar} & 1.9B & 1.65 & 296.5 & 0.83 & 0.60\\
AR & RAR-XXL~\cite{yu2024randomized} & 1.5B & 1.48 & 326.0 & 0.80 & 0.63 \\
\hline
MAR & MAR-B~\cite{mar} & 208M & 2.31 &281.7 &0.82 &0.57 \\
MAR & MAR-L~\cite{mar} &479M& 1.78 &296.0& 0.81& 0.60 \\
MAR & MAR-H~\cite{mar} & 943M&1.55& 303.7& 0.81 &0.62 \\
\hline
VAR&VAR-$d16$~\cite{var}   & 310M  & 3.30& 274.4& 0.84& 0.51    \\
VAR&VAR-$d20$~\cite{var}   &600M & 2.57& 302.6& 0.83& 0.56     \\
VAR&VAR-$d30$~\cite{var}   & 2.0B      & 1.97  & 323.1 & 0.82 & 0.59      \\
\hline
\modelname& \modelname-B    &172M   &1.72&280.4&0.82&0.59 \\
\modelname& \modelname-L   & 608M   & 1.28& 292.5&0.82&0.62\\
\modelname& \modelname-H    & 1.1B    & 1.24 &301.6&0.83&0.64\\
\end{tabular}
\caption{
\textbf{Generation Results on ImageNet-256.}
Metrics include Fréchet Inception Distance (FID), Inception Score (IS), Precision, and Recall. $^\dag$ denotes the use of guidance interval sampling~\cite{guidance}. The proposed \modelname-H achieves a state-of-the-art 1.24 FID on the ImageNet-256 benchmark without relying on vision foundation models (\eg, DINOv2~\cite{dinov2}) or guidance interval sampling~\cite{guidance}, as used in REPA~\cite{yu2024representation}.
}\label{tab:256}
\end{table*}

\subsection{Main Results}
\label{sec:main}
We conduct experiments on ImageNet~\cite{deng2009imagenet} at 256$\times$256 and 512$\times$512 resolutions. Following prior works~\cite{dit,mar}, we evaluate model performance using FID~\cite{fid}, Inception Score (IS)~\cite{is}, Precision, and Recall. \modelname is trained with the same hyper-parameters as~\cite{mar,dit} (\eg, 800 training epochs), with model sizes ranging from 172M to 1.1B parameters. See Appendix~\secref{sec:sup_hyper} for hyper-parameter details.

\begin{table}[t]
    \centering
    \begin{tabular}{c|c|c|c}
      model    &  \#params & FID$\downarrow$ & IS$\uparrow$ \\
      \shline
      VQGAN~\cite{vqgan}&227M &26.52& 66.8\\
      BigGAN~\cite{biggan}& 158M&8.43 &177.9\\
      MaskGiT~\cite{maskgit}& 227M&7.32& 156.0\\
      DiT-XL/2~\cite{dit} &675M &3.04& 240.8 \\
     DiMR-XL/3R~\cite{liu2024alleviating}& 525M&2.89 &289.8 \\
     VAR-d36~\cite{var}  & 2.3B& 2.63 & 303.2\\
     REPA$^\ddagger$~\cite{yu2024representation}&675M &2.08& 274.6 \\
     \hline
     \modelname-L & 608M&1.70& 281.5 \\
    \end{tabular}
    \caption{
    \textbf{Generation Results on ImageNet-512.} $^\ddagger$ denotes the use of DINOv2~\cite{dinov2}.
    }
    \label{tab:512}
\end{table}

\noindent\textbf{ImageNet-256.}
In~\tabref{tab:256}, we compare \modelname with previous state-of-the-art generative models.
Out best variant, \modelname-H, achieves a new state-of-the-art-performance of 1.24 FID, outperforming the GAN-based StyleGAN-XL~\cite{stylegan-xl} by 1.06 FID, masked-prediction-based MaskBit~\cite{maskgit} by 0.28 FID, AR-based RAR~\cite{yu2024randomized} by 0.24 FID, VAR~\cite{var} by 0.73 FID, MAR~\cite{mar} by 0.31 FID, and flow-matching-based REPA~\cite{yu2024representation} by 0.18 FID.
Notably, \modelname does not rely on vision foundation models~\cite{dinov2} or guidance interval sampling~\cite{guidance}, both of which were used in REPA~\cite{yu2024representation}, the previous best-performing model.
Additionally, our lightweight \modelname-B (172M), surpasses DiT-XL (675M)~\cite{dit} by 0.55 FID while achieving an inference speed of 9.8 images per second—20$\times$ faster than DiT-XL (0.5 images per second). Detailed speed comparison can be found in Appendix \ref{sec:speed}.

\noindent\textbf{ImageNet-512.}
In~\tabref{tab:512}, we report the performance of \modelname on ImageNet-512.
Similarly, \modelname-L sets a new state-of-the-art FID of 1.70, outperforming the diffusion based DiT-XL/2~\cite{dit} and DiMR-XL/3R~\cite{liu2024alleviating} by a large margin of 1.34 and 1.19 FID, respectively.
Additionally, \modelname-L also surpasses the previous best autoregressive model VAR-d36~\cite{var} and flow-matching-based REPA~\cite{yu2024representation} by 0.93 and 0.38 FID, respectively.

\begin{figure*}
    \centering
    \vspace{-6pt}
    \includegraphics[width=1\linewidth]{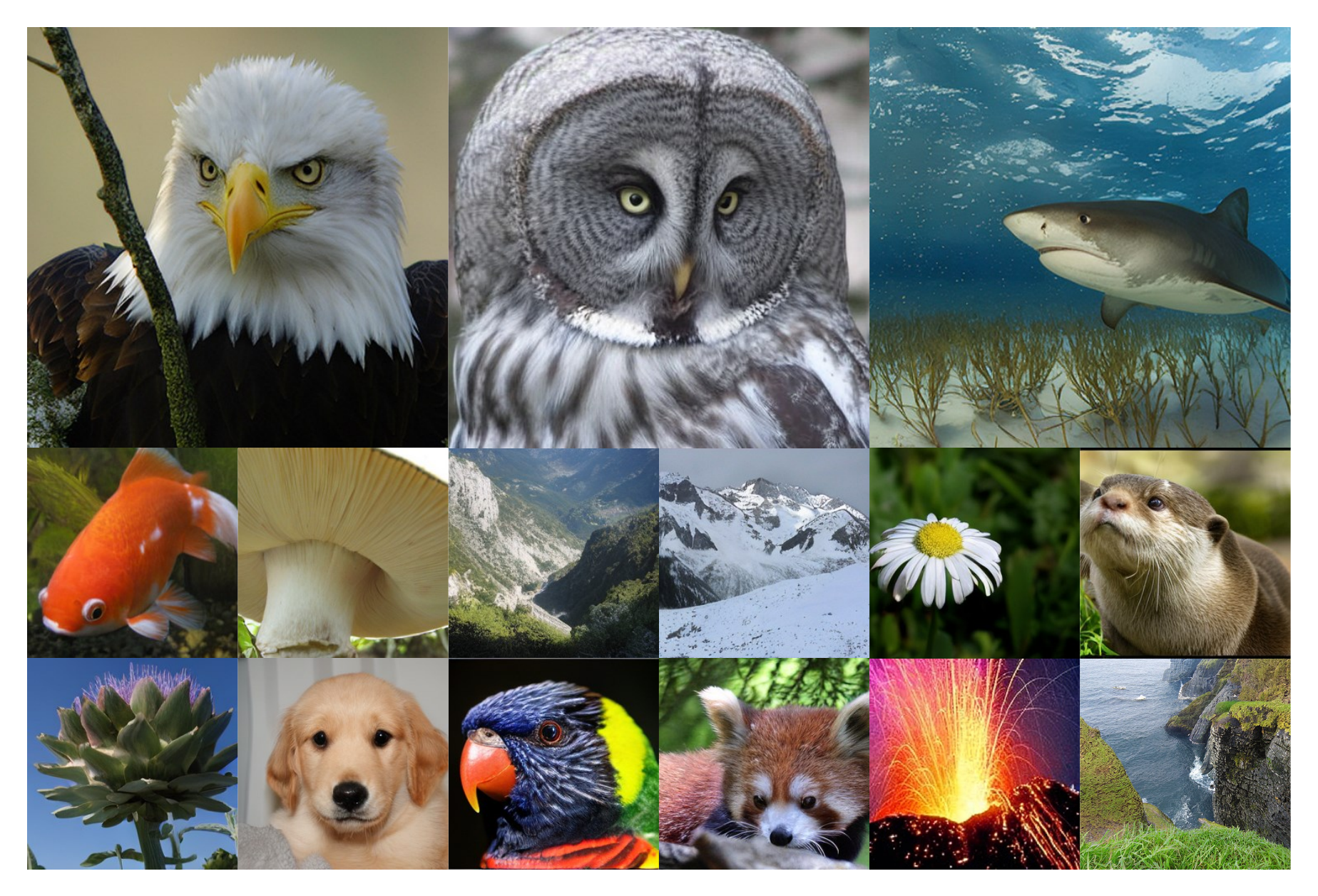}
    \caption{\textbf{Generated Samples.} \modelname generates high-quality images at resolutions of 512$\times$512 (1st row) and 256$\times$256 (2nd and 3rd row).
    }
    \label{fig:qualitative}
\end{figure*}

\noindent\textbf{Qualitative Results.}
\figref{fig:qualitative} presents samples generated by \modelname (trained on ImageNet) at 512$\times$512 and 256$\times$256 resolutions. These results highlight \modelname's ability to produce high-fidelity images with exceptional visual quality.

\subsection{Ablation Studies}
\label{sec:ablation}
In this section, we conduct ablation studies using \modelname-B, trained for 400 epochs to efficiently iterate on model design.

\noindent\textbf{Prediction Entity X.}
The proposed \modelname extends next-token prediction to next-X prediction. In~\tabref{tab:X}, we evaluate different designs for the prediction entity X, including an individual patch token, a cell (a group of surrounding tokens), a subsample (a non-local grouping), a scale (coarse-to-fine resolution), and an entire image.

Among these variants, cell-based \modelname achieves the best performance, with an FID of 2.48, outperforming the token-based \modelname by 1.03 FID and surpassing the second best design (scale-based \modelname) by 0.42 FID. Furthermore, even when using standard prediction entities such as tokens, subsamples, images, or scales, \modelname consistently outperforms existing methods while requiring significantly fewer parameters. These results highlight the efficiency and effectiveness of \modelname across diverse prediction entities.

\begin{table}[]
    \centering
    \scalebox{0.92}{
    \begin{tabular}{c|c|c|c|c}
        model & \makecell[c]{prediction\\entity} & \#params & FID$\downarrow$ & IS$\uparrow$\\
        \shline
        LlamaGen-L~\cite{llamagen} & \multirow{2}{*}{token} & 343M & 3.80 &248.3\\
        \modelname-B& & 172M&3.51&251.4\\
        \hline
        PAR-L~\cite{par} & \multirow{2}{*}{subsample}& 343M & 3.76 & 218.9\\
        \modelname-B&  &172M& 3.58&231.5\\
        \hline
        DiT-L/2~\cite{dit}& \multirow{2}{*}{image}& 458M&5.02&167.2 \\
         \modelname-B& & 172M&3.13&253.4 \\
        \hline
        VAR-$d16$~\cite{var} & \multirow{2}{*}{scale} & 310M&3.30 &274.4\\
        \modelname-B& &172M&2.90&262.8\\
        \hline
        \baseline{\modelname-B}& \baseline{cell} & \baseline{172M}&\baseline{2.48}&\baseline{269.2} \\
    \end{tabular}
    }
    \caption{\textbf{Ablation on Prediction Entity X.} Using cells as the prediction entity outperforms alternatives such as tokens or entire images. Additionally, under the same prediction entity, \modelname surpasses previous methods, demonstrating its effectiveness across different prediction granularities. }%
    \label{tab:X}
\end{table}

\noindent\textbf{Cell Size.}
A prediction entity cell is formed by grouping spatially adjacent $k\times k$ tokens, where a larger cell size incorporates more tokens and thus captures a broader context within a single prediction step.
For a $256\times256$ input image, the encoded continuous latent representation has a spatial resolution of $16\times16$. Given this, the image can be partitioned into an $m\times m$ grid, where each cell consists of $k\times k$ neighboring tokens. As shown in~\tabref{tab:cell}, we evaluate different cell sizes with $k \in \{1,2,4,8,16\}$, where $k=1$ represents a single token and $k=16$ corresponds to the entire image as a single entity. We observe that performance improves as $k$ increases, peaking at an FID of 2.48 when using cell size $8\times8$ (\ie, $k=8$). Beyond this, performance declines, reaching an FID of 3.13 when the entire image is treated as a single entity.
These results suggest that using cells rather than the entire image as the prediction unit allows the model to condition on previously generated context, improving confidence in predictions while maintaining both rich semantics and local details.

\begin{table}[t]
    \centering
    \scalebox{0.98}{
    \begin{tabular}{c|c|c|c}
    cell size ($k\times k$ tokens) & $m\times m$ grid & FID$\downarrow$ & IS$\uparrow$ \\
       \shline
       $1\times1$ & $16\times16$ &3.51&251.4 \\
       $2\times2$ & $8\times8$ & 3.04& 253.5\\
       $4\times4$ & $4\times4$ & 2.61&258.2 \\
       \baseline{$8\times8$} & \baseline{$2\times2$} & \baseline{2.48} & \baseline{269.2}\\
       $16\times16$ & $1\times1$ & 3.13&253.4  \\
    \end{tabular}
    }
    \caption{\textbf{Ablation on the cell size.}
    In this study, a $16\times16$ continuous latent representation is partitioned into an $m\times m$ grid, where each cell consits of $k\times k$ neighboring tokens.
    A cell size of $8\times8$ achieves the best performance, striking an optimal balance between local structure and global context.
    }
    \label{tab:cell}
\end{table}

\begin{table}[t]
    \centering
    \scalebox{0.95}{
    \begin{tabular}{c|c|c|c}
      previous cell & noise time step &  FID$\downarrow$ & IS$\uparrow$ \\
       \shline
       clean & $t_i=0, \forall i<n$& 3.45& 243.5\\
       increasing noise & $t_1<t_2<\cdots<t_{n-1}$& 2.95&258.8 \\
       decreasing noise & $t_1>t_2>\cdots>t_{n-1}$&2.78 &262.1 \\
      \baseline{random noise}  & \baseline{no constraint} &\baseline{2.48} & \baseline{269.2}\\
    \end{tabular}
    }
    \caption{
    \textbf{Ablation on Noisy Context Learning.}
    This study examines the impact of noise time steps ($t_1, \cdots, t_{n-1} \subset [0, 1]$) in previous entities ($t=0$ represents pure Gaussian noise).
    Conditioning on all clean entities (the ``clean'' variant) results in suboptimal performance.
    Imposing an order on noise time steps, either ``increasing noise'' or ``decreasing noise'', also leads to inferior results. The best performance is achieved with the "random noise" setting, where no constraints are imposed on noise time steps.
    }
    \label{tab:ncl}
\end{table}

\noindent\textbf{Noisy Context Learning.}
During training, \modelname employs Noisy Context Learning (NCL), predicting $X_n$ by conditioning on all previous noisy entities, unlike Teacher Forcing.
The noise intensity of previous entities is contorlled by noise time steps $\{t_1, \dots, t_{n-1}\} \subset [0, 1]$, where $t=0$ corresponds to pure Gaussian noise.
We analyze the impact of NCL in~\tabref{tab:ncl}.
When conditioning on all clean entities (\ie, the ``clean'' variant, where $t_i=0, \forall i<n$), which is equivalent to vanilla AR (\ie, Teacher Forcing), the suboptimal performance is obtained.
We also evaluate two constrained noise schedules: the ``increasing noise'' variant, where noise time steps increase over AR steps ($t_1<t_2< \cdots < t_{n-1}$), and the `` decreasing noise'' variant, where noise time steps decrease ($t_1>t_2> \cdots > t_{n-1}$).
While both settings improve over the ``clean'' variant, they remain inferior to our final ``random noise'' setting, where no constraints are imposed on noise time steps, leading to the best performance.

\section{Conclusion}
In this work, we introduced \modelname, a general next-X prediction framework for autoregressive visual generation.
Unlike traditional next-token prediction, \modelname reformulates discrete token classification as continuous entity regression, enabling more flexible and semantically meaningful prediction units.
Through systematic exploration, we found that next-cell prediction provides the best balance between local structure and global coherence.
To mitigate exposure bias, we proposed Noisy Context Learning (NCL), which trains the model on noisy entities instead of pristine ground truth inputs, improving robustness and reducing cascading errors.
As a result, \modelname achieves state-of-the-art performance on ImageNet-256 and ImageNet-512.

{
    \small
    \bibliographystyle{ieeenat_fullname}
    \bibliography{main}

\begin{thebibliography}{63}
\providecommand{\natexlab}[1]{#1}
\providecommand{\url}[1]{\texttt{#1}}
\expandafter\ifx\csname urlstyle\endcsname\relax
  \providecommand{\doi}[1]{doi: #1}\else
  \providecommand{\doi}{doi: \begingroup \urlstyle{rm}\Url}\fi

\bibitem[dit(2024)]{dit-github}
Alpha-vllm. large-dit-imagenet.
\newblock 2024.

\bibitem[Arora et~al.(2022)Arora, El~Asri, Bahuleyan, and Cheung]{acc1}
Kushal Arora, Layla El~Asri, Hareesh Bahuleyan, and Jackie Chi~Kit Cheung.
\newblock Why exposure bias matters: An imitation learning perspective of error accumulation in language generation.
\newblock In \emph{Findings of the Association for Computational Linguistics: ACL 2022}, 2022.

\bibitem[Bao et~al.(2023)Bao, Nie, Xue, Cao, Li, Su, and Zhu]{uvit}
Fan Bao, Shen Nie, Kaiwen Xue, Yue Cao, Chongxuan Li, Hang Su, and Jun Zhu.
\newblock All are worth words: A vit backbone for diffusion models.
\newblock In \emph{CVPR}, 2023.

\bibitem[Brock et~al.(2018)Brock, Donahue, and Simonyan]{biggan}
Andrew Brock, Jeff Donahue, and Karen Simonyan.
\newblock Large scale gan training for high fidelity natural image synthesis.
\newblock \emph{arXiv preprint arXiv:1809.11096}, 2018.

\bibitem[Brown et~al.(2020)Brown, Mann, Ryder, Subbiah, Kaplan, Dhariwal, Neelakantan, Shyam, Sastry, Askell, et~al.]{gpt3}
Tom~B. Brown, Benjamin Mann, Nick Ryder, Melanie Subbiah, Jared Kaplan, Prafulla Dhariwal, Arvind Neelakantan, Pranav Shyam, Girish Sastry, Amanda Askell, et~al.
\newblock Language models are few-shot learners.
\newblock \emph{NeurIPS}, 2020.

\bibitem[Chang et~al.(2022)Chang, Zhang, Jiang, Liu, and Freeman]{maskgit}
Huiwen Chang, Han Zhang, Lu Jiang, Ce Liu, and William~T Freeman.
\newblock Maskgit: Masked generative image transformer.
\newblock In \emph{CVPR}, 2022.

\bibitem[Chen et~al.(2020)Chen, Radford, Child, Wu, Jun, Dhariwal, Luan, and Sutskever]{igpt}
Mark Chen, Alec Radford, Rewon Child, Jeff Wu, Heewoo Jun, Prafulla Dhariwal, David Luan, and Ilya Sutskever.
\newblock Generative pretraining from pixels.
\newblock In \emph{ICML}, 2020.

\bibitem[Deng et~al.(2024)Deng, Zh, Li, Guan, and Fan]{deng2024causal}
Chaorui Deng, Deyao Zh, Kunchang Li, Shi Guan, and Haoqi Fan.
\newblock Causal diffusion transformers for generative modeling.
\newblock \emph{arXiv preprint arXiv:2412.12095}, 2024.

\bibitem[Deng et~al.(2009)Deng, Dong, Socher, Li, Li, and Fei-Fei]{deng2009imagenet}
Jia Deng, Wei Dong, Richard Socher, Li-Jia Li, Kai Li, and Li Fei-Fei.
\newblock Imagenet: A large-scale hierarchical image database.
\newblock In \emph{CVPR}, 2009.

\bibitem[Dhariwal and Nichol(2021)]{adm}
Prafulla Dhariwal and Alexander Nichol.
\newblock Diffusion models beat gans on image synthesis.
\newblock \emph{NeurIPS}, 34, 2021.

\bibitem[Esser et~al.(2021)Esser, Rombach, and Ommer]{vqgan}
Patrick Esser, Robin Rombach, and Bjorn Ommer.
\newblock Taming transformers for high-resolution image synthesis.
\newblock In \emph{CVPR}, 2021.

\bibitem[Esser et~al.(2024)Esser, Kulal, Blattmann, Entezari, M{\"u}ller, Saini, Levi, Lorenz, Sauer, Boesel, et~al.]{sd3}
Patrick Esser, Sumith Kulal, Andreas Blattmann, Rahim Entezari, Jonas M{\"u}ller, Harry Saini, Yam Levi, Dominik Lorenz, Axel Sauer, Frederic Boesel, et~al.
\newblock Scaling rectified flow transformers for high-resolution image synthesis.
\newblock In \emph{ICML}, 2024.

\bibitem[Gao et~al.(2023)Gao, Zhou, Cheng, and Yan]{gao2023mdtv2}
Shanghua Gao, Pan Zhou, Ming-Ming Cheng, and Shuicheng Yan.
\newblock Mdtv2: Masked diffusion transformer is a strong image synthesizer.
\newblock \emph{arXiv preprint arXiv:2303.14389}, 2023.

\bibitem[Goodfellow et~al.(2014)Goodfellow, Pouget-Abadie, Mirza, Xu, Warde-Farley, Ozair, Courville, and Bengio]{gan}
Ian Goodfellow, Jean Pouget-Abadie, Mehdi Mirza, Bing Xu, David Warde-Farley, Sherjil Ozair, Aaron Courville, and Yoshua Bengio.
\newblock Generative adversarial nets.
\newblock \emph{NeurIPS}, 2014.

\bibitem[Gu et~al.(2025)Gu, Wang, Zhang, Zhang, Zhang, Jaitly, Susskind, and Zhai]{gu2025dart}
Jiatao Gu, Yuyang Wang, Yizhe Zhang, Qihang Zhang, Dinghuai Zhang, Navdeep Jaitly, Josh Susskind, and Shuangfei Zhai.
\newblock Dart: Denoising autoregressive transformer for scalable text-to-image generation.
\newblock In \emph{ICLR}, 2025.

\bibitem[He et~al.(2021)He, Zhang, Zhou, and Glass]{acc2}
Tianxing He, Jingzhao Zhang, Zhiming Zhou, and James Glass.
\newblock Exposure bias versus self-recovery: Are distortions really incremental for autoregressive text generation?
\newblock In \emph{EMNLP}, 2021.

\bibitem[Heusel et~al.(2017)Heusel, Ramsauer, Unterthiner, Nessler, and Hochreiter]{fid}
Martin Heusel, Hubert Ramsauer, Thomas Unterthiner, Bernhard Nessler, and Sepp Hochreiter.
\newblock Gans trained by a two time-scale update rule converge to a local nash equilibrium.
\newblock \emph{NeurIPS}, 30, 2017.

\bibitem[Ho et~al.(2020)Ho, Jain, and Abbeel]{diff2}
Jonathan Ho, Ajay Jain, and Pieter Abbeel.
\newblock Denoising diffusion probabilistic models.
\newblock \emph{NeurIPS}, 2020.

\bibitem[Hoogeboom et~al.(2023)Hoogeboom, Heek, and Salimans]{diff1}
Emiel Hoogeboom, Jonathan Heek, and Tim Salimans.
\newblock simple diffusion: End-to-end diffusion for high resolution images.
\newblock In \emph{ICML}, 2023.

\bibitem[Hoogeboom et~al.(2024)Hoogeboom, Mensink, Heek, Lamerigts, Gao, and Salimans]{hoogeboom2024simpler}
Emiel Hoogeboom, Thomas Mensink, Jonathan Heek, Kay Lamerigts, Ruiqi Gao, and Tim Salimans.
\newblock Simpler diffusion (sid2): 1.5 fid on imagenet512 with pixel-space diffusion.
\newblock \emph{arXiv preprint arXiv:2410.19324}, 2024.

\bibitem[Kang et~al.(2023)Kang, Zhu, Zhang, Park, Shechtman, Paris, and Park]{gigagan}
Minguk Kang, Jun-Yan Zhu, Richard Zhang, Jaesik Park, Eli Shechtman, Sylvain Paris, and Taesung Park.
\newblock Scaling up gans for text-to-image synthesis.
\newblock In \emph{CVPR}, 2023.

\bibitem[Kingma and Ba(2015)]{kingma2014adam}
Diederik~P Kingma and Jimmy Ba.
\newblock Adam: A method for stochastic optimization.
\newblock In \emph{ICLR}, 2015.

\bibitem[Kingma and Welling(2014)]{vae}
Diederik~P Kingma and Max Welling.
\newblock Auto-encoding variational bayes.
\newblock In \emph{ICLR}, 2014.

\bibitem[Kynk{\"a}{\"a}nniemi et~al.(2024)Kynk{\"a}{\"a}nniemi, Aittala, Karras, Laine, Aila, and Lehtinen]{guidance}
Tuomas Kynk{\"a}{\"a}nniemi, Miika Aittala, Tero Karras, Samuli Laine, Timo Aila, and Jaakko Lehtinen.
\newblock Applying guidance in a limited interval improves sample and distribution quality in diffusion models.
\newblock \emph{arXiv preprint arXiv:2404.07724}, 2024.

\bibitem[Lee et~al.(2022)Lee, Kim, Kim, Cho, and Han]{rq}
Doyup Lee, Chiheon Kim, Saehoon Kim, Minsu Cho, and Wook-Shin Han.
\newblock Autoregressive image generation using residual quantization.
\newblock In \emph{CVPR}, 2022.

\bibitem[Li et~al.(2024)Li, Tian, Li, Deng, and He]{mar}
Tianhong Li, Yonglong Tian, He Li, Mingyang Deng, and Kaiming He.
\newblock Autoregressive image generation without vector quantization.
\newblock \emph{NeurIPS}, 2024.

\bibitem[Lipman et~al.(2022)Lipman, Chen, Ben-Hamu, Nickel, and Le]{lipman2022flow}
Yaron Lipman, Ricky~TQ Chen, Heli Ben-Hamu, Maximilian Nickel, and Matt Le.
\newblock Flow matching for generative modeling.
\newblock \emph{arXiv preprint arXiv:2210.02747}, 2022.

\bibitem[Liu et~al.(2024)Liu, Zeng, He, Yu, Shen, and Chen]{liu2024alleviating}
Qihao Liu, Zhanpeng Zeng, Ju He, Qihang Yu, Xiaohui Shen, and Liang-Chieh Chen.
\newblock Alleviating distortion in image generation via multi-resolution diffusion models.
\newblock \emph{NeurIPS}, 2024.

\bibitem[Liu et~al.(2022)Liu, Gong, and Liu]{liu2022flow}
Xingchao Liu, Chengyue Gong, and Qiang Liu.
\newblock Flow straight and fast: Learning to generate and transfer data with rectified flow.
\newblock \emph{arXiv preprint arXiv:2209.03003}, 2022.

\bibitem[Loshchilov and Hutter(2019)]{loshchilov2017decoupled}
Ilya Loshchilov and Frank Hutter.
\newblock Decoupled weight decay regularization.
\newblock \emph{ICLR}, 2019.

\bibitem[Luo et~al.(2024)Luo, Shi, Ge, Yang, Wang, and Shan]{luo2024open}
Zhuoyan Luo, Fengyuan Shi, Yixiao Ge, Yujiu Yang, Limin Wang, and Ying Shan.
\newblock Open-magvit2: An open-source project toward democratizing auto-regressive visual generation.
\newblock \emph{arXiv preprint arXiv:2409.04410}, 2024.

\bibitem[Ma et~al.(2024)Ma, Goldstein, Albergo, Boffi, Vanden-Eijnden, and Xie]{sit}
Nanye Ma, Mark Goldstein, Michael~S Albergo, Nicholas~M Boffi, Eric Vanden-Eijnden, and Saining Xie.
\newblock Sit: Exploring flow and diffusion-based generative models with scalable interpolant transformers.
\newblock In \emph{ECCV}, 2024.

\bibitem[OpenAI(2022)]{chatgpt}
OpenAI.
\newblock Introducing chatgpt.
\newblock \url{https://openai.com/blog/chatgpt/}, 2022.

\bibitem[OpenAI(2023)]{gpt4}
OpenAI.
\newblock Gpt-4 technical report.
\newblock \emph{arXiv preprint arXiv:2303.08774}, 2023.

\bibitem[Oquab et~al.(2023)Oquab, Darcet, Moutakanni, Vo, Szafraniec, Khalidov, Fernandez, Haziza, Massa, El-Nouby, et~al.]{dinov2}
Maxime Oquab, Timoth{\'e}e Darcet, Th{\'e}o Moutakanni, Huy Vo, Marc Szafraniec, Vasil Khalidov, Pierre Fernandez, Daniel Haziza, Francisco Massa, Alaaeldin El-Nouby, et~al.
\newblock Dinov2: Learning robust visual features without supervision.
\newblock \emph{arXiv preprint arXiv:2304.07193}, 2023.

\bibitem[Peebles and Xie(2023)]{dit}
William Peebles and Saining Xie.
\newblock Scalable diffusion models with transformers.
\newblock In \emph{ICCV}, 2023.

\bibitem[Radford et~al.(2018)Radford, Narasimhan, Salimans, and Sutskever]{gpt}
Alec Radford, Karthik Narasimhan, Tim Salimans, and Ilya Sutskever.
\newblock Improving language understanding by generative pre-training.
\newblock \url{https://cdn.openai.com/research-covers/language-unsupervised/language_understanding_paper.pdf}, 2018.

\bibitem[Ranzato et~al.(2016)Ranzato, Chopra, Auli, and Zaremba]{ranzato2016sequence}
Marc'Aurelio Ranzato, Sumit Chopra, Michael Auli, and Wojciech Zaremba.
\newblock Sequence level training with recurrent neural networks.
\newblock In \emph{ICLR}, 2016.

\bibitem[Razavi et~al.(2019)Razavi, Van Den~Oord, and Vinyals]{vqvae2}
Ali Razavi, Aaron Van Den~Oord, and Oriol Vinyals.
\newblock Generating diverse high-fidelity images with vq-vae-2.
\newblock \emph{NeurIPS}, 2019.

\bibitem[Ren et~al.(2024)Ren, Yu, He, Shen, Yuille, and Chen]{flowar}
Sucheng Ren, Qihang Yu, Ju He, Xiaohui Shen, Alan Yuille, and Liang-Chieh Chen.
\newblock Flowar: Scale-wise autoregressive image generation meets flow matching.
\newblock \emph{arXiv preprint arXiv:2412.15205}, 2024.

\bibitem[Rombach et~al.(2022)Rombach, Blattmann, Lorenz, Esser, and Ommer]{ldm}
Robin Rombach, Andreas Blattmann, Dominik Lorenz, Patrick Esser, and Bj{\"o}rn Ommer.
\newblock High-resolution image synthesis with latent diffusion models.
\newblock In \emph{CVPR}, 2022.

\bibitem[Ronneberger et~al.(2015)Ronneberger, Fischer, and Brox]{unet}
Olaf Ronneberger, Philipp Fischer, and Thomas Brox.
\newblock U-net: Convolutional networks for biomedical image segmentation.
\newblock In \emph{MICCAI}, 2015.

\bibitem[Salimans et~al.(2016)Salimans, Goodfellow, Zaremba, Cheung, Radford, and Chen]{is}
Tim Salimans, Ian Goodfellow, Wojciech Zaremba, Vicki Cheung, Alec Radford, and Xi Chen.
\newblock Improved techniques for training gans.
\newblock \emph{NeurIPS}, 29, 2016.

\bibitem[Sauer et~al.(2022)Sauer, Schwarz, and Geiger]{stylegan-xl}
Axel Sauer, Katja Schwarz, and Andreas Geiger.
\newblock Stylegan-xl: Scaling stylegan to large diverse datasets.
\newblock \emph{arXiv preprint arXiv:2201.00273}, 2022.

\bibitem[Song et~al.(2020)Song, Meng, and Ermon]{diff3}
Jiaming Song, Chenlin Meng, and Stefano Ermon.
\newblock Denoising diffusion implicit models.
\newblock \emph{arXiv preprint arXiv:2010.02502}, 2020.

\bibitem[Song and Ermon(2019)]{song2019generative}
Yang Song and Stefano Ermon.
\newblock Generative modeling by estimating gradients of the data distribution.
\newblock \emph{NeurIPS}, 2019.

\bibitem[Sun et~al.(2024)Sun, Jiang, Chen, Zhang, Peng, Luo, and Yuan]{llamagen}
Peize Sun, Yi Jiang, Shoufa Chen, Shilong Zhang, Bingyue Peng, Ping Luo, and Zehuan Yuan.
\newblock Autoregressive model beats diffusion: Llama for scalable image generation.
\newblock \emph{arXiv preprint arXiv:2406.06525}, 2024.

\bibitem[Tian et~al.(2024)Tian, Jiang, Yuan, Peng, and Wang]{var}
Keyu Tian, Yi Jiang, Zehuan Yuan, Bingyue Peng, and Liwei Wang.
\newblock Visual autoregressive modeling: Scalable image generation via next-scale prediction.
\newblock \emph{NeurIPS}, 2024.

\bibitem[Touvron et~al.(2023)Touvron, Lavril, Izacard, Martinet, Lachaux, Lacroix, Rozi{\`e}re, Goyal, Hambro, Azhar, et~al.]{llama}
Hugo Touvron, Thibaut Lavril, Gautier Izacard, Xavier Martinet, Marie-Anne Lachaux, Timoth{\'e}e Lacroix, Baptiste Rozi{\`e}re, Naman Goyal, Eric Hambro, Faisal Azhar, et~al.
\newblock Llama: Open and efficient foundation language models.
\newblock \emph{arXiv preprint arXiv:2302.13971}, 2023.

\bibitem[Van Den~Oord et~al.(2016{\natexlab{a}})Van Den~Oord, Kalchbrenner, Espeholt, Vinyals, Graves, et~al.]{pixelcnn}
Aaron Van Den~Oord, Nal Kalchbrenner, Lasse Espeholt, Oriol Vinyals, Alex Graves, et~al.
\newblock Conditional image generation with pixelcnn decoders.
\newblock \emph{NeurIPS}, 2016{\natexlab{a}}.

\bibitem[Van Den~Oord et~al.(2016{\natexlab{b}})Van Den~Oord, Kalchbrenner, and Kavukcuoglu]{pixelrnn}
Aaron Van Den~Oord, Nal Kalchbrenner, and Koray Kavukcuoglu.
\newblock Pixel recurrent neural networks.
\newblock In \emph{ICML}, 2016{\natexlab{b}}.

\bibitem[Van Den~Oord et~al.(2017)Van Den~Oord, Vinyals, et~al.]{vqvae}
Aaron Van Den~Oord, Oriol Vinyals, et~al.
\newblock Neural discrete representation learning.
\newblock \emph{NeurIPS}, 2017.

\bibitem[Vaswani et~al.(2017)Vaswani, Shazeer, Parmar, Uszkoreit, Jones, Gomez, Kaiser, and Polosukhin]{vaswani2017attention}
Ashish Vaswani, Noam Shazeer, Niki Parmar, Jakob Uszkoreit, Llion Jones, Aidan~N Gomez, {\L}ukasz Kaiser, and Illia Polosukhin.
\newblock Attention is all you need.
\newblock \emph{NeurIPS}, 2017.

\bibitem[Wang et~al.(2024)Wang, Ren, Lin, Han, Guo, Yang, Zou, Feng, and Liu]{par}
Yuqing Wang, Shuhuai Ren, Zhijie Lin, Yujin Han, Haoyuan Guo, Zhenheng Yang, Difan Zou, Jiashi Feng, and Xihui Liu.
\newblock Parallelized autoregressive visual generation.
\newblock \emph{arXiv preprint arXiv:2412.15119}, 2024.

\bibitem[Weber et~al.(2024)Weber, Yu, Yu, Deng, Shen, Cremers, and Chen]{weber2024maskbit}
Mark Weber, Lijun Yu, Qihang Yu, Xueqing Deng, Xiaohui Shen, Daniel Cremers, and Liang-Chieh Chen.
\newblock Maskbit: Embedding-free image generation via bit tokens.
\newblock \emph{arXiv preprint arXiv:2409.16211}, 2024.

\bibitem[Williams and Zipser(1989)]{williams1989learning}
Ronald~J Williams and David Zipser.
\newblock A learning algorithm for continually running fully recurrent neural networks.
\newblock \emph{Neural computation}, 1\penalty0 (2):\penalty0 270--280, 1989.

\bibitem[Yu et~al.(2021)Yu, Li, Koh, Zhang, Pang, Qin, Ku, Xu, Baldridge, and Wu]{vit-vqgan}
Jiahui Yu, Xin Li, Jing~Yu Koh, Han Zhang, Ruoming Pang, James Qin, Alexander Ku, Yuanzhong Xu, Jason Baldridge, and Yonghui Wu.
\newblock Vector-quantized image modeling with improved vqgan.
\newblock \emph{arXiv preprint arXiv:2110.04627}, 2021.

\bibitem[Yu et~al.(2022)Yu, Xu, Koh, Luong, Baid, Wang, Vasudevan, Ku, Yang, Ayan, et~al.]{parti}
Jiahui Yu, Yuanzhong Xu, Jing~Yu Koh, Thang Luong, Gunjan Baid, Zirui Wang, Vijay Vasudevan, Alexander Ku, Yinfei Yang, Burcu~Karagol Ayan, et~al.
\newblock Scaling autoregressive models for content-rich text-to-image generation.
\newblock \emph{arXiv preprint arXiv:2206.10789}, 2022.

\bibitem[Yu et~al.(2024{\natexlab{a}})Yu, Lezama, Gundavarapu, Versari, Sohn, Minnen, Cheng, Gupta, Gu, Hauptmann, et~al.]{yu2024language}
Lijun Yu, Jos{\'e} Lezama, Nitesh~B Gundavarapu, Luca Versari, Kihyuk Sohn, David Minnen, Yong Cheng, Agrim Gupta, Xiuye Gu, Alexander~G Hauptmann, et~al.
\newblock Language model beats diffusion--tokenizer is key to visual generation.
\newblock In \emph{ICLR}, 2024{\natexlab{a}}.

\bibitem[Yu et~al.(2024{\natexlab{b}})Yu, He, Deng, Shen, and Chen]{yu2024randomized}
Qihang Yu, Ju He, Xueqing Deng, Xiaohui Shen, and Liang-Chieh Chen.
\newblock Randomized autoregressive visual generation.
\newblock \emph{arXiv preprint arXiv:2411.00776}, 2024{\natexlab{b}}.

\bibitem[Yu et~al.(2024{\natexlab{c}})Yu, Weber, Deng, Shen, Cremers, and Chen]{yu2024image}
Qihang Yu, Mark Weber, Xueqing Deng, Xiaohui Shen, Daniel Cremers, and Liang-Chieh Chen.
\newblock An image is worth 32 tokens for reconstruction and generation.
\newblock \emph{NeurIPS}, 2024{\natexlab{c}}.

\bibitem[Yu et~al.(2024{\natexlab{d}})Yu, Kwak, Jang, Jeong, Huang, Shin, and Xie]{yu2024representation}
Sihyun Yu, Sangkyung Kwak, Huiwon Jang, Jongheon Jeong, Jonathan Huang, Jinwoo Shin, and Saining Xie.
\newblock Representation alignment for generation: Training diffusion transformers is easier than you think.
\newblock \emph{arXiv preprint arXiv:2410.06940}, 2024{\natexlab{d}}.

\bibitem[Zhao et~al.(2024)Zhao, Song, Wang, Feng, Ding, Sun, Xiao, and Wang]{zhao2024monoformer}
Chuyang Zhao, Yuxing Song, Wenhao Wang, Haocheng Feng, Errui Ding, Yifan Sun, Xinyan Xiao, and Jingdong Wang.
\newblock Monoformer: One transformer for both diffusion and autoregression.
\newblock \emph{arXiv preprint arXiv:2409.16280}, 2024.

\end{thebibliography}
}

\clearpage
\setcounter{page}{1}
\maketitlesupplementary

\renewcommand{\thesection}{\Alph{section}}
\setcounter{section}{0}

\section*{Appendix}
\label{sec:appendix}

The supplementary material includes the following additional information:

\begin{itemize}
    \item \secref{sec:sup_hyper} details the hyper-parameters used for \modelname.
    \item \secref{sec:speed} provides a comprehensive speed comparison.
    \item \secref{sec:limitation} discusses the limitations and future directions.
    \item \secref{sec:sup_vis} presents visualization samples generated by \modelname.
\end{itemize}

\section{Hyper-parameters for \modelname}
\label{sec:sup_hyper}
We list the detailed training and inference hyper-parameters in~\tabref{tab:hparams}.

\begin{table}[h!]
\centering

\tablestyle{5.0pt}{1.1}
\begin{tabular}{l|c}
config \quad\quad\quad\quad\quad\quad\quad\quad & value \\
\hline
optimizer & AdamW~\cite{kingma2014adam,loshchilov2017decoupled} \\
optimizer momentum & (0.9, 0.96) \\
weight decay & 0.02 \\
batch size & 2048 \\
learning rate schedule & cosine decay \\
peak learning rate & 4e-4 \\
ending learning rate & 1e-5 \\
total epochs & 800 \\
warmup epochs & 100 \\
dropout rate & 0.1 \\
attn dropout rate & 0.1 \\
class label dropout rate & 0.1 \\
inference mode & SDE \\
inference steps & 50 \\
\end{tabular}
\caption{\textbf{Detailed Hyper-parameters of \modelname Models.}
}
\label{tab:hparams}
\end{table}

\section{Speed Comparison.}
We compare \modelname with diffusion-, flow matching-, and autoregressive-based models in~\tabref{tab:sampling_speed}. Our most lightweight variant, \modelname-B (172M), outperforms DiT-XL (diffusion-based), SiT-XL (flow matching-based), and MAR (autoregressive-based), while achieving a 20$\times$ speedup (9.8 \vs \ 0.5 images/sec).
Additionally, \modelname-L surpasses the recent state-of-the-art model REPA, running 5.3$\times$ faster (3.2 \vs \ 0.6 images/sec).
Finally, our largest model, \modelname-H, achieves 1.24 FID on ImageNet-256, setting a new state-of-the-art, while still running 2.2$\times$ faster than REPA.

\label{sec:speed}
\begin{table}
\centering
\tablestyle{3.0pt}{1.05}
\begin{tabular}{c|ccccc}
method & type & \#params & FID$\downarrow$ & steps & images/sec \\
\shline
DiT-XL/2~\cite{dit} & Diff. & 675M & 2.27 & 250 & 0.5 \\
SiT-XL/2~\cite{sit} & Flow. & 675M & 2.02 & 250 & 0.5 \\
MAR-L~\cite{mar} & AR & 479M & 1.78 & 256 & 0.5 \\
\modelname-B &xAR&172M&1.72 &50&9.8\\
\hline
MAR-H~\cite{mar} & MAR & 943M & 1.55 & 256 & 0.3 \\
REPA~\cite{yu2024representation} & Flow. &675M& 1.42 & 250 & 0.6 \\
\modelname-L &xAR&608M&1.28 &50&3.2 \\
\hline
\modelname-H &xAR&1.1B&1.24&50&1.3 \\
\end{tabular}
\caption{\textbf{Sampling Throughput Comparison.} Throughputs are evaluated as samples generated per second on a single A100 based on their official codebases.
}
\label{tab:sampling_speed}
\end{table}

\section{Discussion and Limitations}
\label{sec:limitation}
Our empirical evaluations indicate that a square 8$\times$8 cell configuration achieves the best performance, with no noticeable difference when using rectangular cells (\eg, $k/2\times 2k$ or $2k\times k/2$), which introduce additional complexity without clear benefits.
Given that different regions in an image contain varying levels of semantic information (\eg, dense object areas \vs \ uniform sky regions), future research could explore whether dynamically shaped prediction entities provide additional benefits. However, in this work, we adopt a simple yet effective square cell design, demonstrating state-of-the-art results on the challenging ImageNet generation benchmark.

\section{Visualization of Generated Samples}
\label{sec:sup_vis}

\begin{figure}[!h]
    \centering
    \includegraphics[width=\linewidth]{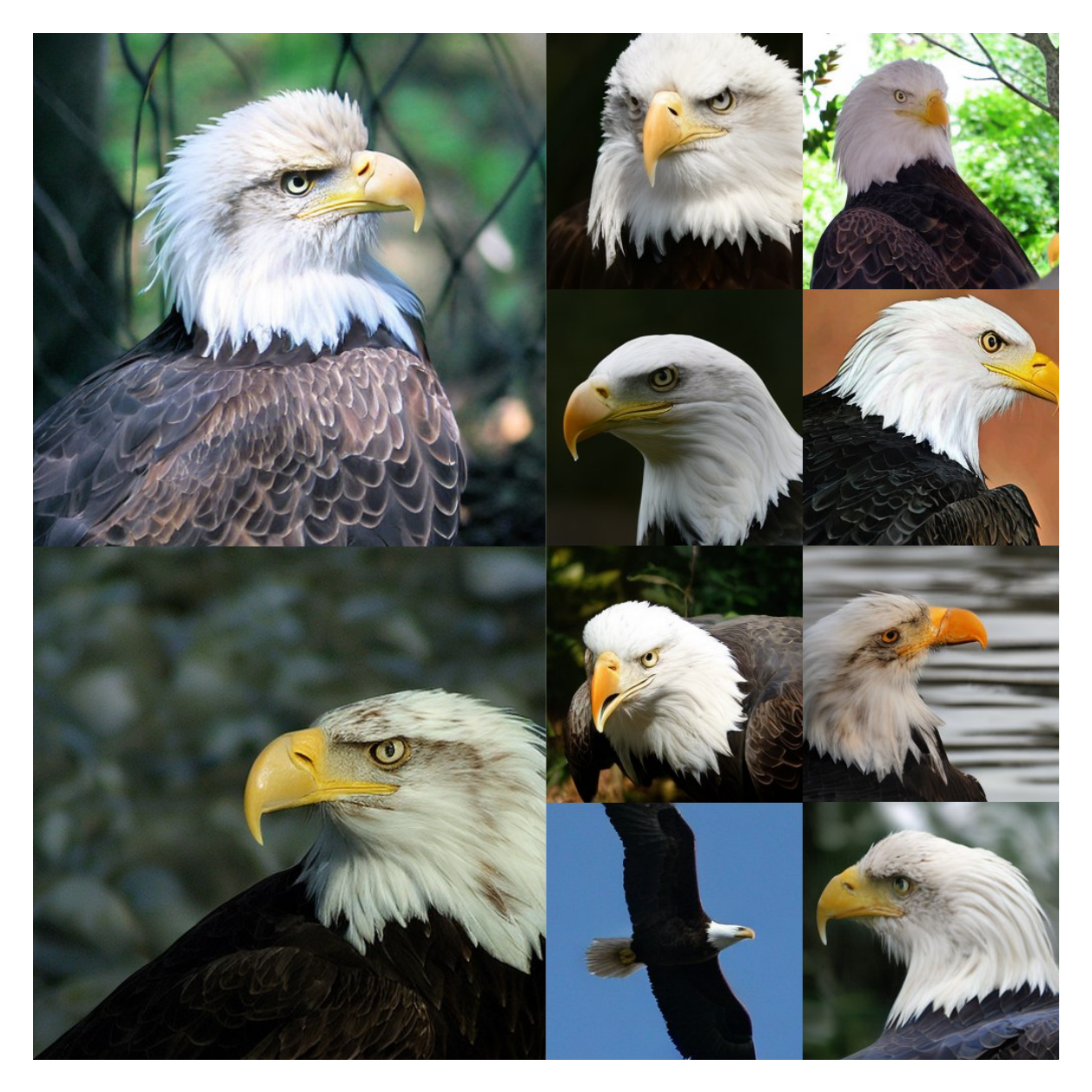}
    \caption{\textbf{Generated Samples from \modelname.} \modelname is able to generate high-fidelity American eagle (22) images.}
    \label{fig:22}
\end{figure}

\begin{figure}
    \centering
    \includegraphics[width=\linewidth]{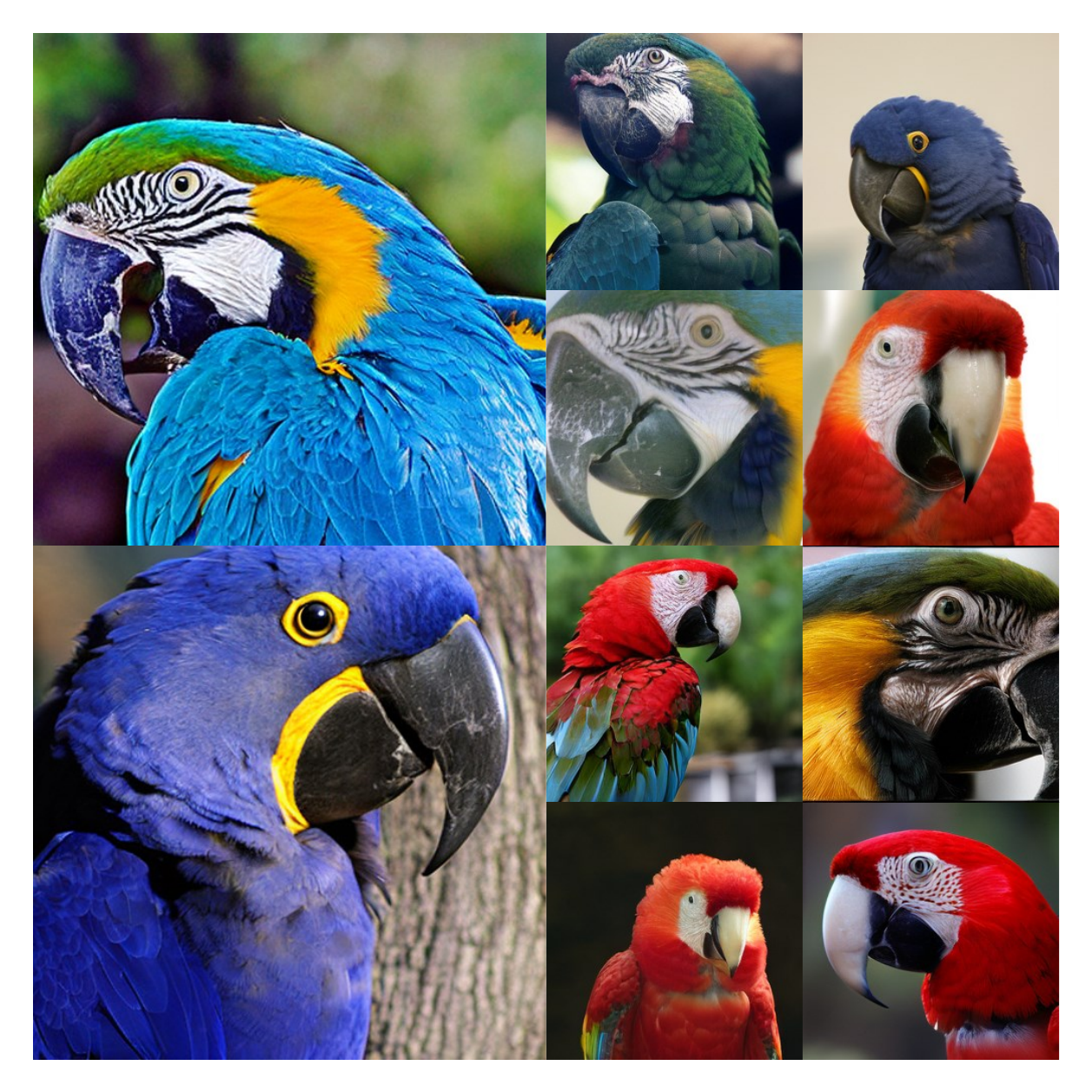}
    \caption{\textbf{Generated Samples from \modelname.} \modelname is able to generate high-fidelity macaw (88) images.
    }
    \label{fig:88}
\end{figure}

\begin{figure}
    \centering
    \includegraphics[width=\linewidth]{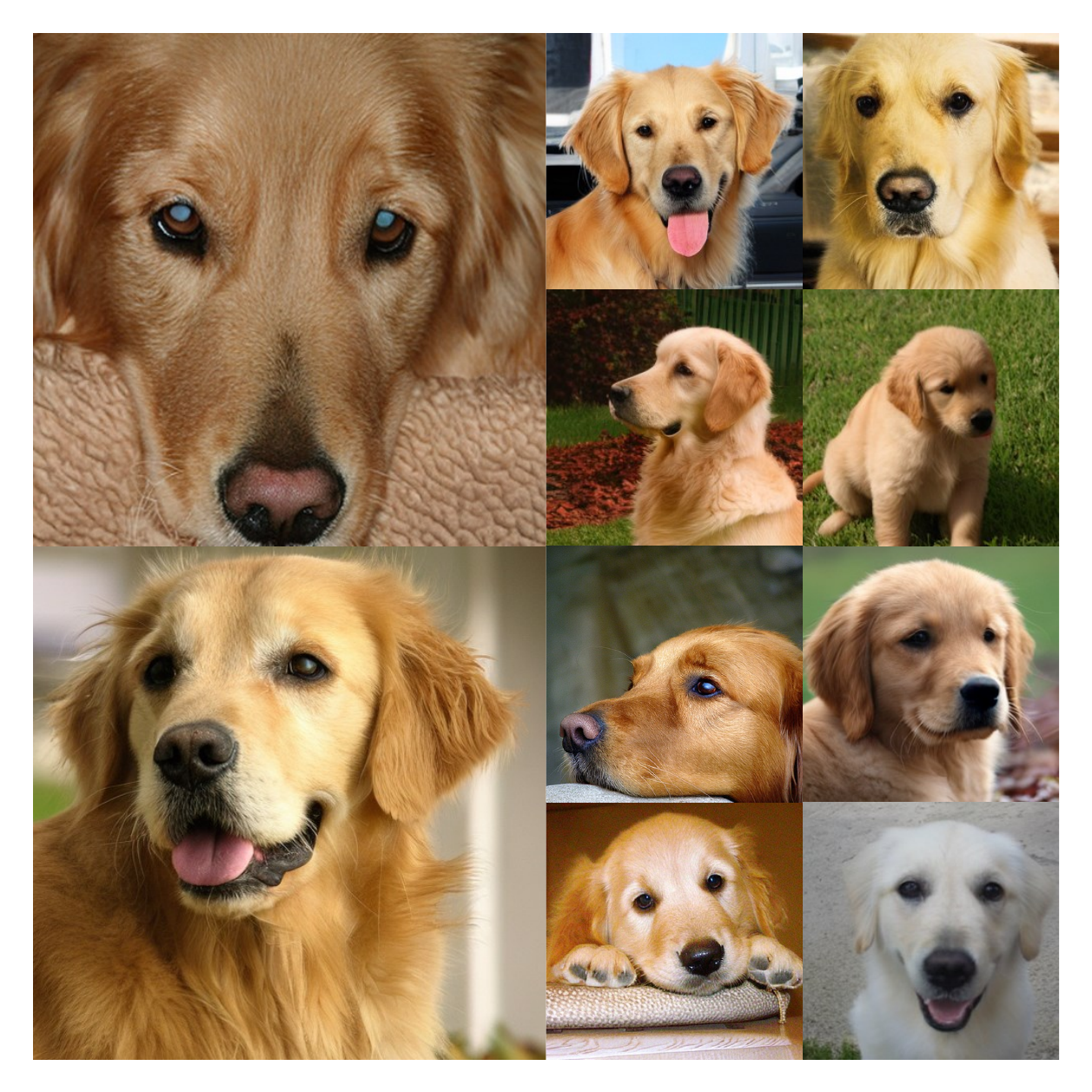}
    \caption{\textbf{Generated Samples from \modelname.} \modelname is able to generate high-fidelity golden retriever (207) images.}
    \label{fig:207}
\end{figure}

\begin{figure}
    \centering
    \includegraphics[width=\linewidth]{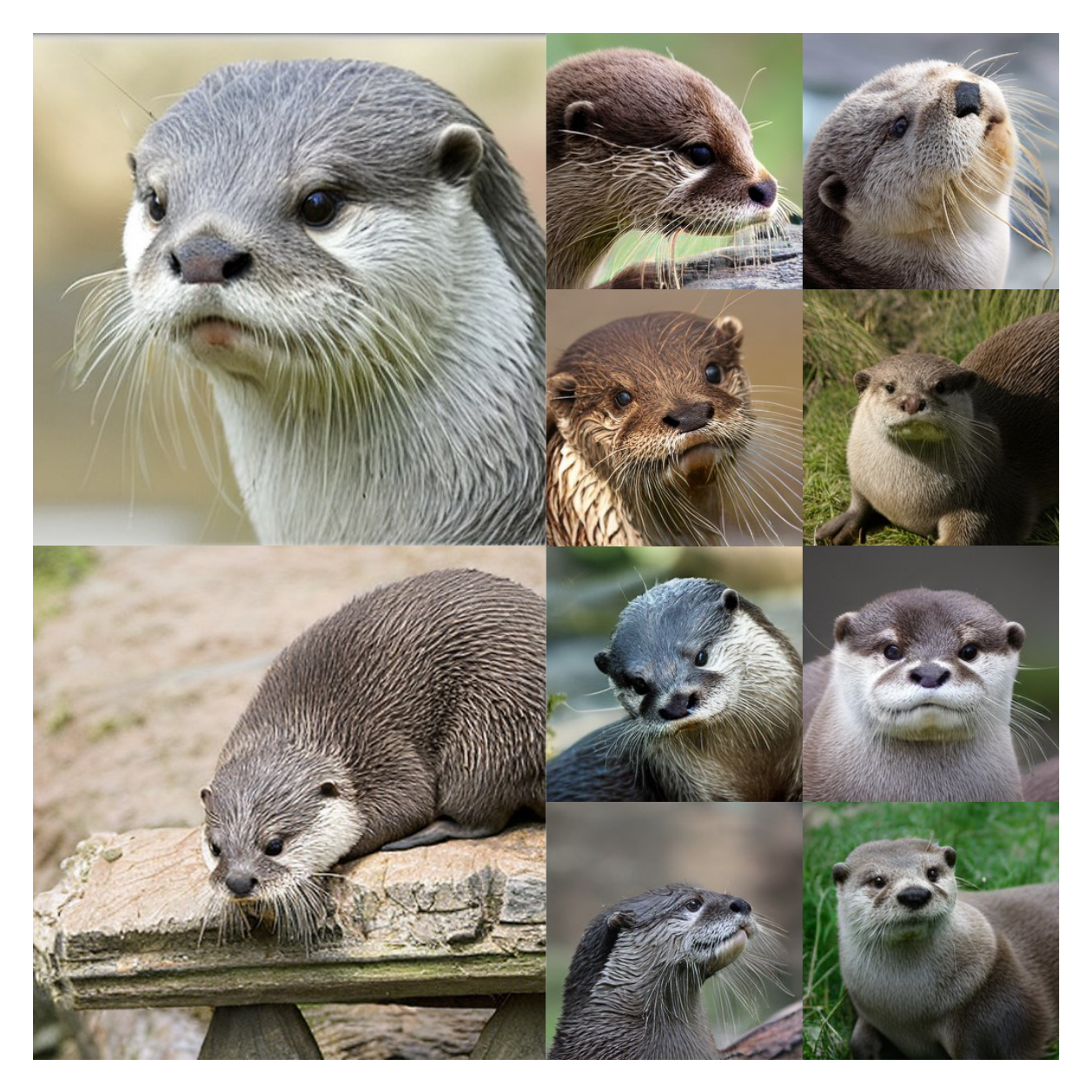}
    \caption{\textbf{Generated Samples from \modelname.} \modelname is able to generate high-fidelity otter (360) images.}
    \label{fig:360}
\end{figure}

\begin{figure}
    \centering
    \includegraphics[width=\linewidth]{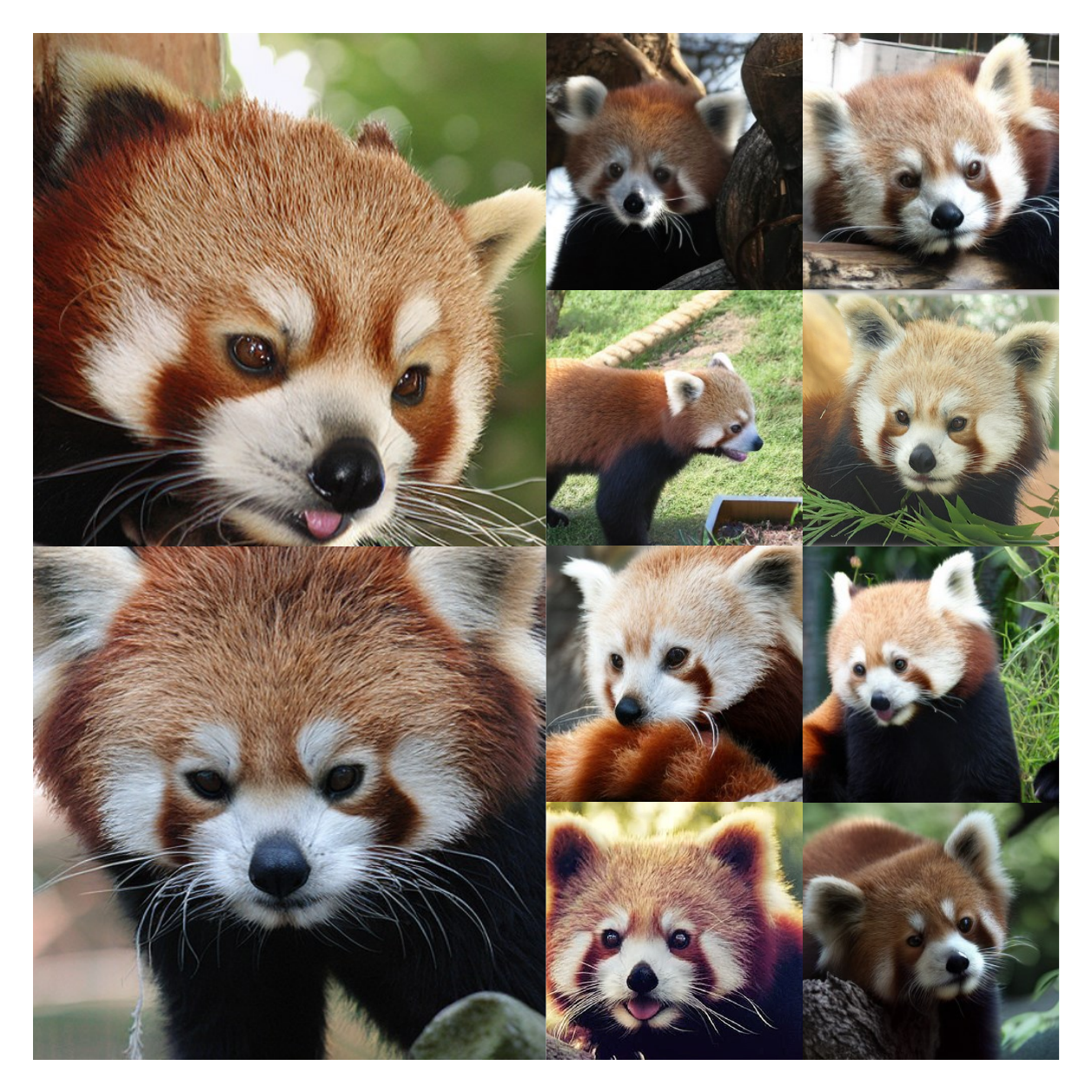}
    \caption{\textbf{Generated Samples from \modelname.} \modelname is able to generate high-fidelity lesser panda (387) images.}
    \label{fig:387}
\end{figure}

\begin{figure}
    \centering
    \includegraphics[width=\linewidth]{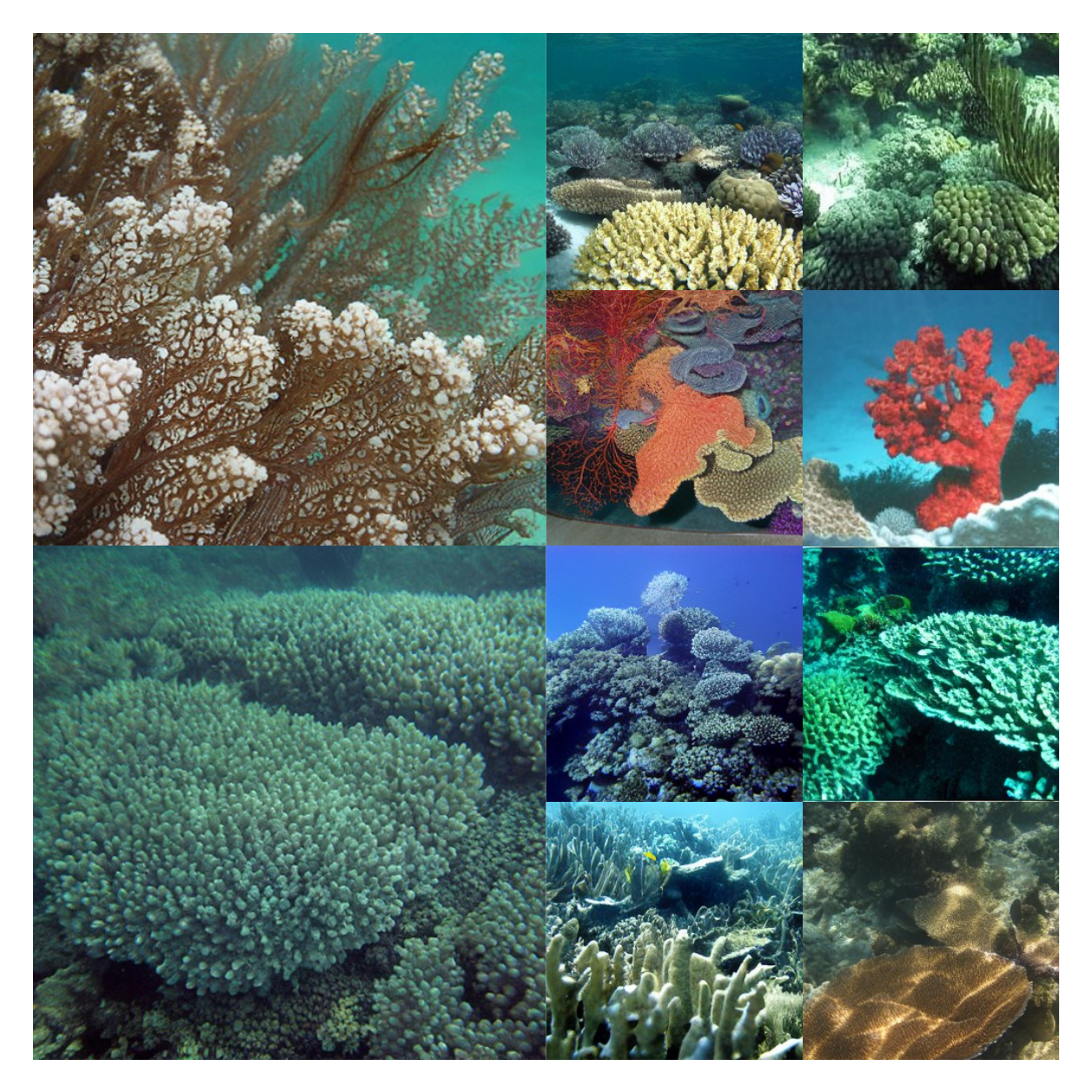}
    \caption{\textbf{Generated Samples from \modelname.} \modelname is able to generate high-fidelity coral reef (973) images.}
    \label{fig:973}
\end{figure}

\begin{figure}
    \centering
    \includegraphics[width=\linewidth]{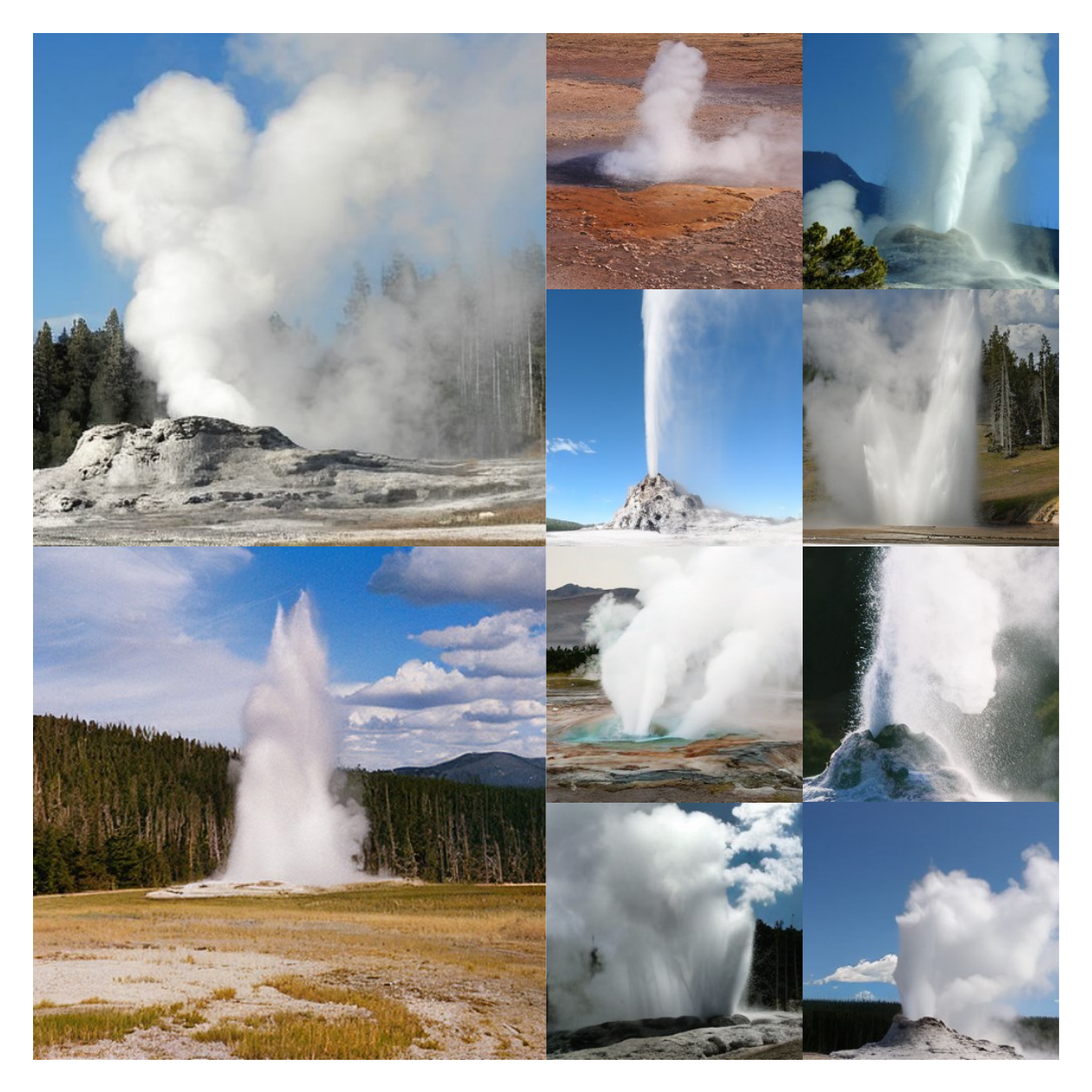}
    \caption{\textbf{Generated Samples from \modelname.} \modelname is able to generate high-fidelity geyser (974) images.}
    \label{fig:974}
\end{figure}

\begin{figure}
    \centering
    \includegraphics[width=\linewidth]{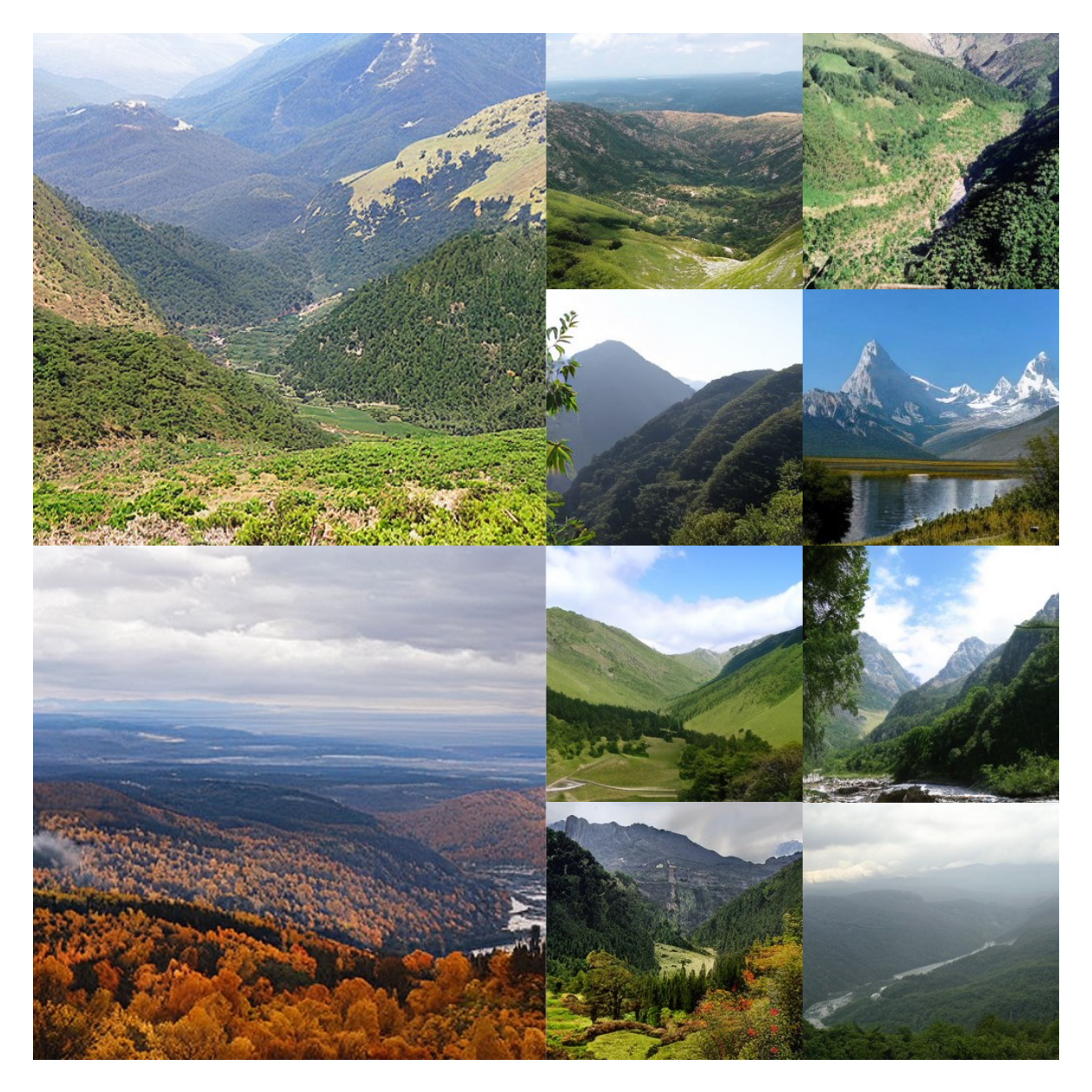}
    \caption{\textbf{Generated Samples from \modelname.} \modelname is able to generate high-fidelity valley (979) images.}
    \label{fig:979}
\end{figure}

\begin{figure}
    \centering
    \includegraphics[width=\linewidth]{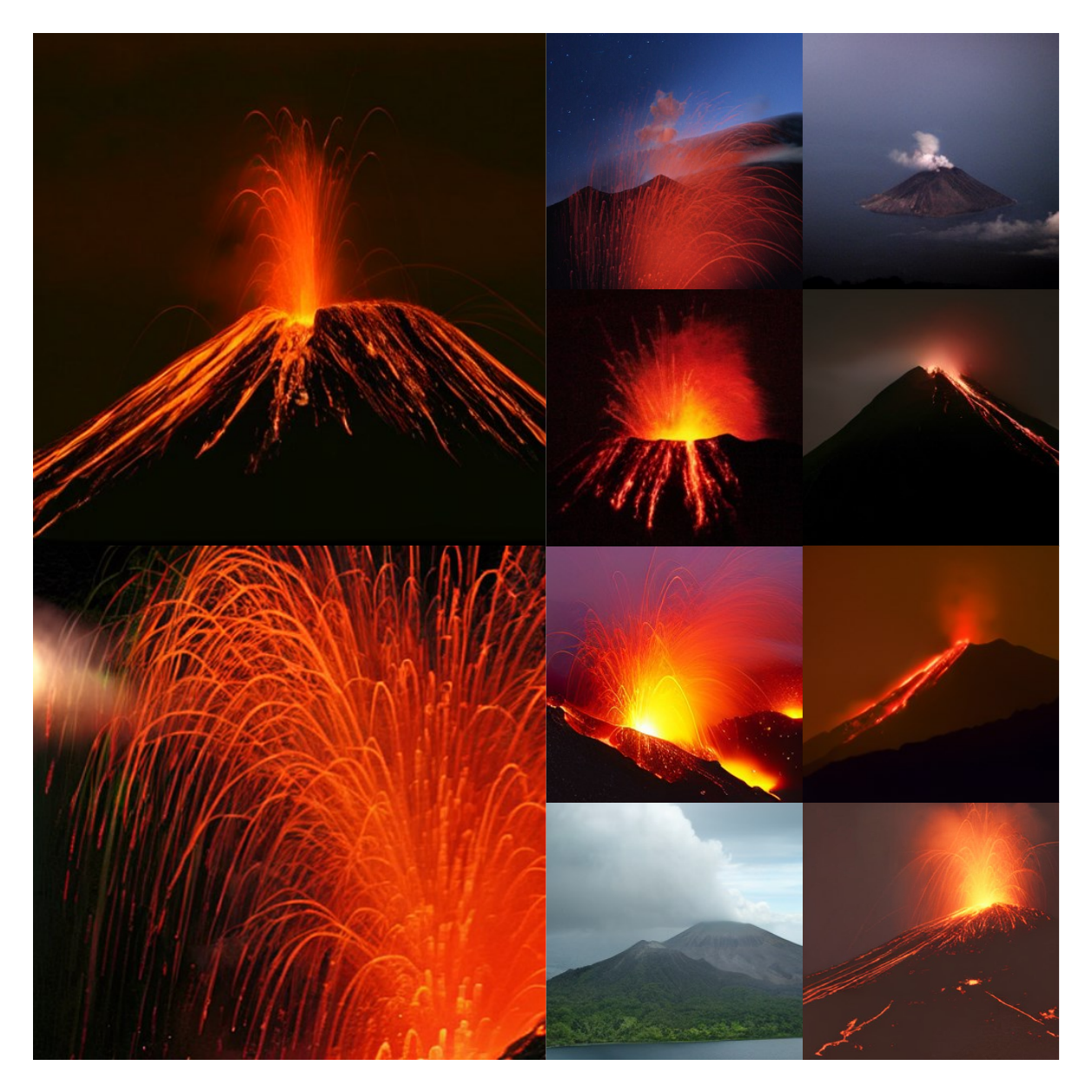}
    \caption{\textbf{Generated Samples from \modelname.} \modelname is able to generate high-fidelity volcano (980) images.}
    \label{fig:980}
\end{figure}

Additional visualization results generated by \modelname-H are provided from~\figref{fig:22} to~\figref{fig:980}.

\end{document}